\documentclass[a4paper,fleqn]{cas-dc}

\usepackage[authoryear]{natbib}
\usepackage{hyperref}
\usepackage{graphicx}
\usepackage{subfigure}
\usepackage{epstopdf}
\usepackage{xcolor}
\usepackage[noend]{algpseudocode}
\usepackage{algorithmicx,algorithm}
\usepackage{color}
\usepackage{amsmath}
\usepackage{amsfonts}
\usepackage{algorithm}
\usepackage{algpseudocode}
\usepackage{graphicx}
\usepackage{subcaption}
\usepackage{amsmath}
\usepackage{enumitem}
\usepackage{multirow}
\usepackage{diagbox}
\usepackage{booktabs}
\usepackage{enumitem}
\usepackage[justification=centering]{caption}
\newtheorem{myexample}{Example}
\newtheorem{mytheorem}{Theorem}
\newtheorem{mydefinition}{Definition}
\newtheorem{myLemma}{Lemma}
\def\tsc#1{\csdef{#1}{\textsc{\lowercase{#1}}\xspace}}
\tsc{WGM}
\tsc{QE}
\tsc{EP}
\tsc{PMS}
\tsc{BEC}
\tsc{DE}

\begin{document}
\begin{sloppypar}
	\let\WriteBookmarks\relax
	\def\floatpagepagefraction{1}
	\def\textpagefraction{.001}
	\let\printorcid\relax
	\shorttitle{}
	\shortauthors{R. Miao et~al.} 

	\title [mode = title]{ParMod: A Parallel and Modular Framework for Learning Non-Markovian Tasks}



	\author[1]{\textcolor[RGB]{0,0,1}{Ruixuan Miao}}
	\ead{22031212106@stu.xidian.edu.cn}

	\author[1]{\textcolor[RGB]{0,0,1}{Xu Lu}}
	\address[1]{Institute of Computing Theory and Technology and State Key Laboratory of ISN, Xidian University, PR China}
	\cormark[1]
	\ead{xlu@xidian.edu.cn}

	\author[1]{\textcolor[RGB]{0,0,1}{Cong Tian}}
\cormark[1]
\ead{ctian@mail.xidian.edu.cn}

    \author[1]{\textcolor[RGB]{0,0,1}{Bin Yu}}
        \cormark[1]
	\ead{byu@xidian.edu.cn}
    
    \author[1]{\textcolor[RGB]{0,0,1}{Zhenhua Duan}}
	\ead{zhhduan@mail.xidian.edu.cn}

	\cortext[cor1]{Corresponding author, joint first author.} 



	\begin{abstract}
        The commonly used Reinforcement Learning (RL) model, MDPs (Markov Decision Processes), has a basic premise that rewards depend on the current state and action only. However, many real-world tasks are non-Markovian, which has long-term memory and dependency. The reward sparseness problem is further amplified in non-Markovian scenarios. Hence learning a non-Markovian task (NMT) is inherently more difficult than learning a Markovian one. In this paper, we propose a novel \textbf{Par}allel and \textbf{Mod}ular RL framework, ParMod, specifically for learning NMTs specified by temporal logic. With the aid of formal techniques, the NMT is modulaized into a series of sub-tasks based on the automaton structure (equivalent to its temporal logic counterpart). On this basis, sub-tasks will be trained by a group of agents in a parallel fashion, with one agent handling one sub-task. Besides parallel training, the core of ParMod lies in: a flexible classification method for modularizing the NMT, and an effective reward shaping method for improving the sample efficiency. A comprehensive evaluation is conducted on several challenging benchmark problems with respect to various metrics. The experimental results show that ParMod achieves superior performance over other relevant studies. Our work thus provides a good synergy among RL, NMT and temporal logic.
            
        \end{abstract}
		

	\maketitle

\section{Introduction}
\label{introduction}

Reinforcement Learning (RL) is a well-known paradigm for autonomous decision-making in complex and unknown environments \cite{sutton2018reinforcement}. It is used to help agents learn a policy, actually a strategy in pursuit of goals, from trial-and-error experiences towards maximizing long-term rewards. RL, or deep RL, achieves remarkable success in a wide range of tasks, from games \cite{silver2016mastering} to real-world context \cite{kalashnikov2018qt}. An important mathematical foundation for RL is Markov Decision Processes (MDPs). In MDPs, the accumulation of rewards relies only on the current state and action. However, there are many real-world tasks whose rewards are in response to more complex behaviors that is reflected over historical states and actions. Such rewards are often known as non-Markovian rewards, and tasks with these rewards are called non-Markovian tasks (NMTs). For instance, an autonomous taxi will receive reward for picking up passengers and subsequently delivering them to their destinations. An RL agent that attempts to learn NMTs without realizing that they are non-Markovian will display sub-optimal behaviors.

Learning on NMTs has gained a lot of attractions recently, and a number of different solution methods are explored. NMTs are common in real-world applications, while state-of-the-art RL algorithms have limited success on them due to the non-Markovian feature. We address the challenge of ensuring non-Markovian requirements in RL from a formal methods perspective. To this end, we adopt temporal logic as an unambiguous specification language of what non-Markovian mean.

In this paper, we propose a novel RL framework for parallel training NMTs, introducing a task modularization unit into the traditional RL setting. Sub-tasks are computed upfront from the given task specification by the unit. In addition, it is essential to design a suitable and dense reward, especially for NMTs. To deal with this challenging problem, our work devises an automatic reward shaping technique to guide the agent towards the fulfilment of task specification.

The remainder of the paper starts with a detailed discussion of related work. Section 3 gives a review of the basic concepts and notations used in this paper. Section 4 elaborates our approach, mainly in how to decompose a non-Markovian task and how to train the task in parallel. Section 5 demonstrate the optimality and correctness of our approach. In Section 5, we provide experimental evaluations on benchmark problems to show the performance of our approach. Finally, we draw conclusions and puts forward the future research directions. All of our codes and data are open-sourced and provided in the GitHub repository\footnote{https://github.com/syemichel/ParMod/}.

\section{Related work}


\subsection{Solutions for Non-Markovian Tasks}

The work \cite{camacho2018non} presents a means of specifying non-Markovian rewards, expressed in LTL$_f$ (Linear Temporal Logic over finite traces) \cite{de2013linear}. Non-Markovian reward functions are put into automata representations, and reshaped automata-based rewards are exploited by off-the-shelf MDP planners to guide search. This approach is quite limited since it cannot be extended to continuous domains. A more general form of NMTs is proposed in \cite{icarte2022reward}, called reward machines which are inherently a special kind of finite state automaton. Different methodologies based on Q-learning and hierarchical RL, to cooperate with reward machines, are presented, including automated reward shaping, task decomposition, and counterfactual reasoning for data augmentation. The authors find a primary drawback that reward shaping in their approach does not help in continuous domains. Focusing on specifying NMTs with multiple sub-objectives and safety constraints, \cite{jothimurugan2019composable} propose a simpler language for formally specifying NMTs and an algorithm called SPECTRL to learn policies of NMTs. Based on quantitative semantics of their specification language, SPECTRL is able to automatically generate reward function and reshape rewards. Although experiments demonstrate that SPECTRL can outperform the state-of-the-art baselines at the time, the complex reward shaping mechanism makes it difficult to extend to other RL algorithms.

\cite{hasanbeig2020deep} investigate a deep RL method for policy synthesis in continuous-state/action unknown environments under requirements expressed in LTL. An LTL specification is converted to a Limit Deterministic B{\"{u}}chi Automaton (LDBA) and synchronised on-the-fly with the agent/environment. A modular Deep Deterministic Policy Gradient (DDPG) architecture is proposed to generate a low-level control policy that maximizes the probability of the given LTL specification. The synchronisation process automatically modularizes a complex global task into sub-tasks in order to deal with the sparse reward problem.
\cite{voloshin2022policy} study the problem of policy optimization with LTL constraints. A learning algorithm is derived based on a reduction from the product of MDP and LDBA to a reachability problem. This approach enjoys a sample complexity analysis for guaranteeing task satisfaction and cost optimality. However, continuous state and action spaces are beyond the capabilities of this approach.
An RL framework is presented in \cite{bozkurt2020control} to synthesise a control policy from a product of LDBA and MDPs. The primary contribution of the framework is to introduce a novel rewarding and discounting scheme based on the B{\"{u}}chi acceptance condition. Motion planning case studies solved by a basic Q-learning implementation are given to show the optimization of the satisfaction probability of the B{\"{u}}chi objective.

By studying the satisfaction of temporal properties on unknown stochastic processes with continuous state space, \cite{kazemi2020formal} propose a sequential learning procedure on the basis of a path-dependent reward function. The positive standard form of the LTL specification forms the reward function which is maximized via the learning procedure. 
\cite{oura2020reinforcement} present a method for the synthesis of a policy satisfying a control specification described by an LTL formula, which is converted to an augmented Limit-Deterministic Generalized B{\"{u}}chi Automaton (LDGBA). A product of augmented LDGBA and MDP is generated, based on which a reward function is defined to relax the sparsity of rewards. Moreover, LTL is considered as strict and hard constraints to ensure safe RL 
\cite{DBLP:conf/nips/KondaT99, COHEN2023101295}.

\subsection{Distributed Reinforcement Learning}

\cite{nair2015massively} present the first massively distributed architecture Gorila for deep RL, which distributes DQN (Deep Q-Network) \cite{mnih2013playing} agents across multiple machines, and uses asynchronous SGD for a centralized Q function learning. The distribution of learning process is intensively studied in A3C \cite{mnih2016asynchronous}, which introduces a conceptually simple and lightweight framework, enabling multiple actor-learners compute gradients and update shared parameters asynchronously. Inspired by A3C, \cite{espeholt2018impala} develop a highly scalable distributed framework IMPALA, which scales to a substantial number of machines without sacrificing data efficiency or resource utilisation. IMPALA achieves better performance than previous works with less data, and crucially exhibits positive transfer between tasks as a result of its multi-task approach. 
\cite{heess2017emergence} explore how a rich environment help to promote the learning of complex behavior. The authors propose DPPO, a distributed version of PPO (Proximal Policy Optimization) \cite{schulman2017proximal}, for training various locomotion skills in a diverse set of environments. Ape-X is a value-based distributed architecture similar to Gorila \cite{horgan2018distributed}. It aims to reduce variance and accelerate convergence, and benefits from importance sampling and prioritized experience replay. Built upon Ape-X, R2D2 adapts recurrent experience replay for RNN-based DQN agents \cite{kapturowski2018recurrent}. \cite{li2023parallel} propose a Parallel Q-Learning scheme for massively parallel GPU-based simulation, including data collection, policy function learning, and value function learning on GPU. Based on a novel abstraction of RL training data flows, a scalable, efficient, and extensible distributed RL system ReaLly Scalable RL is developed \cite{mei2023srl}.

\section{Preliminaries}
 

\subsection{Linear Temporal Logic over Finite Traces}
LTL$_f$ is a variant of LTL concentrating on expressing temporal properties over finite traces. Let $AP$ be a set of atomic propositions. The syntax of LTL$_f$ is defined as follows:
 \begin{equation}
\varphi ::= p \mid \neg \varphi \mid \varphi_1\wedge \varphi_2 \mid \bigcirc \varphi \mid \varphi_1 \text{U} \varphi_2\nonumber
\end{equation}
where $p\in AP$ is an atomic proposition, $\bigcirc$(next) and $\text{U}$(until) are temporal operators. Propositional binary connectives $\vee, \rightarrow, \leftrightarrow$ and boolean values $\top$, $\bot$ can be derived in terms of basic operators. Other temporal operators can also be expressed. For example, $\Diamond$(eventually) and $\Box$(always) are defined by: $\Diamond \varphi\equiv \top \text{U} \varphi, \Box\varphi\equiv\neg\Diamond\neg\varphi$. $\bigcirc\varphi$ denotes that $\varphi$ holds in the next state which must exist.
 $\varphi_1 \text{U} \varphi_2$ indicates that $\varphi_1$ holds until $\varphi_2$ is true. $\Diamond\varphi$ means that $\varphi$ will eventually hold before or right in the last state. $\Box\varphi$ represents that $\varphi$ holds along the whole trace. 
 
A state $s$ is a subset of $AP$ that is true, while atomic propositions in $AP \setminus s$ are assumed to be false. LTL$_f$ formulae are interpreted over finite traces of states $\sigma=s_0...s_n$. The semantics of LTL$_f$ is defined as follows. We say that $\sigma $ satisfies $\varphi$, written as $\sigma \models \varphi $, when $\sigma,0 \models \varphi$.
\begin{itemize}
[label=\textbullet]
    \item $\sigma,i\models p$\quad \textbf{iff} \quad $p \in s_i$.
    \item $\sigma,i \models \neg \varphi$ \quad \textbf{iff} \quad $\sigma,i \not\models \varphi$.
    \item $\sigma,i\models \varphi_1 \wedge \varphi_2$ \quad \textbf{iff} \quad $\sigma,i \models \varphi_1$ and $\sigma,i\models \varphi_2$.
    \item $\sigma,i \models \bigcirc \varphi$ \quad \textbf{iff} \quad $i<n$ and  $\sigma,(i+1)\models \varphi$.
    \item $\sigma,i\models\varphi_1 \text{U} \varphi_2$ \quad \textbf{iff} \quad there exists $i\le j\le n$ such that $ \sigma,j\models\varphi_2$, and $\sigma,k \models \varphi_1$ for each $i\le k < j$.
\end {itemize}

Theoretically, each LTL$_f$ formula can be transformed to a Deterministic Finite Automaton (DFA) that recognizes the same language \cite{de2021compositional}. A DFA is a tuple $\mathcal{A} =\langle Q,\Sigma,\delta,q_0,F \rangle$, where $Q$ is a finite set of states, $\Sigma$ is a finite set of input alphabet, $\delta:Q\times\Sigma\to Q$ is a transition function, $q_0\in Q$ is an initial state, $F\subseteq Q$ is a set of accepting states. Let $\sigma =s_1...s_n$ be a \textit{state trace} over the alphabet $\Sigma=2^{AP}$. We say that $\mathcal{A} $ accepts $\sigma$ (i.e., $\sigma$ satisfies $\varphi$) if there exists a \textit{DFA trace} $\rho = q_0...q_n$ such that $q_{i+1}=\delta(q_{i}, s_{i+1})$ for $0\le i < n$ and $q_n\in F$. $\sigma$ ($\rho$) is called an accepting state trace (DFA trace). The states $E \subseteq Q$ that can never reach $F$ are regarded as the \textit{error states}.

To make it more clear, we distinguish $s$ a state and $q$ a DFA state in the sequel. In this work, LTL$_f$ is employed to specify NMTs.

\begin{figure}[h]
\centering
\includegraphics[width=0.25\textwidth]{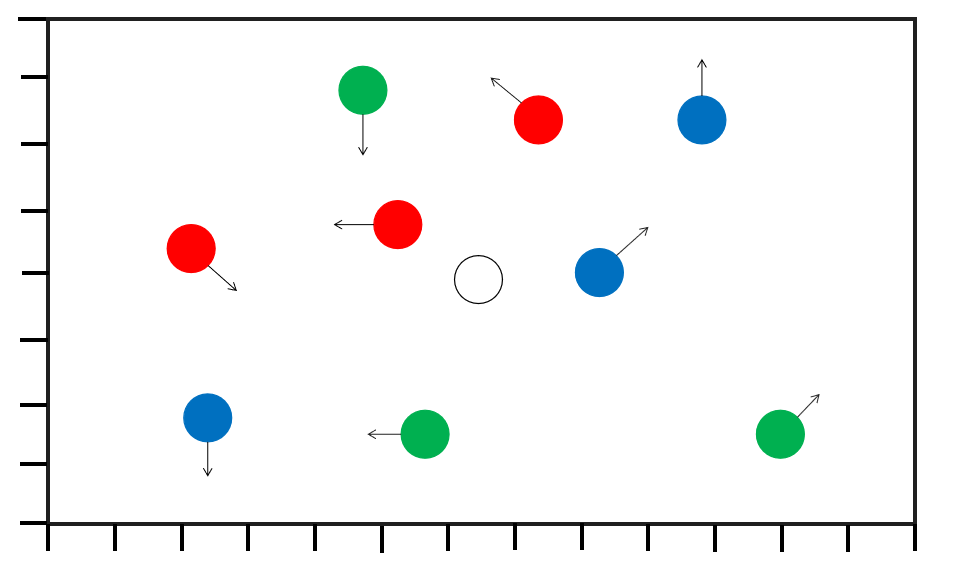}
\caption{An instance of the \emph{Waterworld} problem} 
\label{waterworld}
\end{figure}

\begin{myexample}[\emph{Waterworld} problem - detailed]\label{ex: control knowledge}
     The environment comprises a two-dimensional container with balls of diverse colors inside. Each ball travels in one direction at a constant speed and rebounds when it hits the boundary. The agent, depicted as a white ball, is capable of accelerating or decelerating in any direction.

As shown in Figure \ref{waterworld}, there are three kinds of colored balls. The goal is to touch those balls in a specific order: first touch a red ball and then a green one. This is a typical NMT which can be described by the LTL$_f$ formula $\varphi$ below, where propositions $r$ and $g$ mean touching a red and a green ball respectively. $\varphi$ restricts that the agent must touch a red ball at first. After that, it cannot touch a red ball again, and must touch one of the green ones.
\begin{equation}
\varphi = \left ( \neg r \wedge \neg g \right ) \emph{U}((r \wedge \neg g) \wedge  \bigcirc ((\neg r \wedge  \neg g)\emph{U} (g \wedge \neg r)))  \nonumber
\end{equation}
Note that whether to touch a blue ball is irrelevant to the satisfication of $\varphi$. The corresponding DFA is shown in Figure \ref{DFA}. Obviously, $q2$ is an error state, while $q4$ (double circled) represents an accepting state. The task is considered completed (or failed) when $q4$ (or $q2$) is reached.

\end{myexample}

\subsection{MDP and NMRDP}
  As one of the theoretical foundations of RL, an MDP $M = \langle S, A, R, P, \gamma, s_{0} \rangle$ serves as a model for an agent’s sequential decision-making process. Here, $S$ is a set of states, $A$ is a set of actions, $R:S \times A\times S \longrightarrow \mathbb{R}$ is a reward function, $P(s_{t+1}|s_t, a_t) \in [0,1]$ is a transition probability distribution over the set of next states, given that the agent takes action $a_t$ in state $s_t$ at step $t$ and reaches state $s_{t+1}$, $\gamma$ is the discount factor, $s_0 \in S$ is the initial state.

MDPs are limited in expressiveness due to its memoryless feature, while Non-Markovian Reward Decision Processes (NMRDPs) are more powerful by extending MDPs with non-Markovian rewards, which is suitable for modeling NMTs \cite{thiebaux2006decision}. An NMRDP is a tuple $\text{\emph{NM}}= \langle S, A, R, P, \gamma, s_{0} \rangle$, where $S, A, P, \gamma, s_0$ are the same as those in an MDP. The only difference is the reward function $R$, which is defined as $(S\times A)^*\to \mathbb{R}$. This function indicates that non-Markovian rewards are determined by a finite sequence of states and actions (memory). Consequently, LTL$_f$ is a perfect match with non-Markovian rewards, which is defined in terms of a pair $R = \langle \varphi, r \rangle$. $\varphi$ is an LTL$_f$ formula (reward formula) and $r \in \mathbb{R}$ is the associated reward.

Since the standard training model for RL is based on MDP, advanced RL algorithms cannot be directly used to learn NMTs modeled by NMRDP. To overcome this problem, analogous to priori researches, we apply a synchronisation technique to generate a product of an LTL$_f$ formula (in fact a DFA) and NMRDP in a on-the-fly fashion. Given an NMRDP $\text{\emph{NM}}= \langle S, A, R=\langle \varphi, r \rangle, P, \gamma, s_{0} \rangle$ and a DFA $\mathcal{A} =\langle Q,\Sigma,\delta,q_0,F \rangle$ with respect to the reward formula $\varphi$, the converted MDP is a product $M^\otimes =\left \langle S^\otimes,A,R^\otimes,P^\otimes,\gamma,s^\otimes_0\right \rangle$, where $S^\otimes=S\times Q$ and $s^\otimes_0=\left \langle s_0,q_0  \right \rangle$. Given $s^\otimes= \langle s,q \rangle \in S^\otimes$, $s^\otimes$ is referred to as a \textit{product state}, $s$ is the \textit{state ingredient} of $s^\otimes$, and $q$ is the \textit{DFA ingredient} of $s^\otimes$. The reward function is defined as: 
\begin{equation}
     R^\otimes(s^\otimes_t, a_t, s^\otimes_{t+1})=\left\{\begin{matrix}
  r, & s^\otimes_{t+1}=\left \langle s_{t+1},q_{t+1} \right \rangle, q_{t+1}\in F\\
 0,& \text{otherwise}
\end{matrix}\right.\nonumber
\end{equation}
which means that the agent receives a reward $r$ only when the DFA ingredient of $s_{t+1}^\otimes$ is an accepting state. The transition probability $P^\otimes(s_{t+1}^\otimes\mid s_t^\otimes,a_t)$ equals to $P(s_{t+1}\mid s_t,a_t)$ if there exists a possible DFA transition enabled by $s_t$ from the DFA ingredient of $s_{t}^\otimes$ to that of $s_{t+1}^\otimes$, and 0 otherwise, which is formalized below:
\begin{flushleft}
  $P^\otimes(s_{t+1}^\otimes\mid s_t^\otimes,a_t) = \nonumber$
\end{flushleft}
\begin{flushleft}
  $\left\{\begin{matrix}
  P(s_{t+1}\mid s_t,a_t),& s_{t}^\otimes = \langle s_{t}, q_{t} \rangle, s_{t+1}^\otimes = \langle s_{t+1}, q_{t+1} \rangle, \\ &\exists \nu \in \Sigma : \left \langle q_t,\nu ,q_{t+1} \right \rangle \in \delta \text{ and } s_t\models \nu \\
  0,& \text{otherwise}
\end{matrix}\right.\nonumber$
\end{flushleft}


\section{Parallel and Modular RL Framework}

\begin{figure}[t]
\centering
\includegraphics[width=0.48\textwidth]{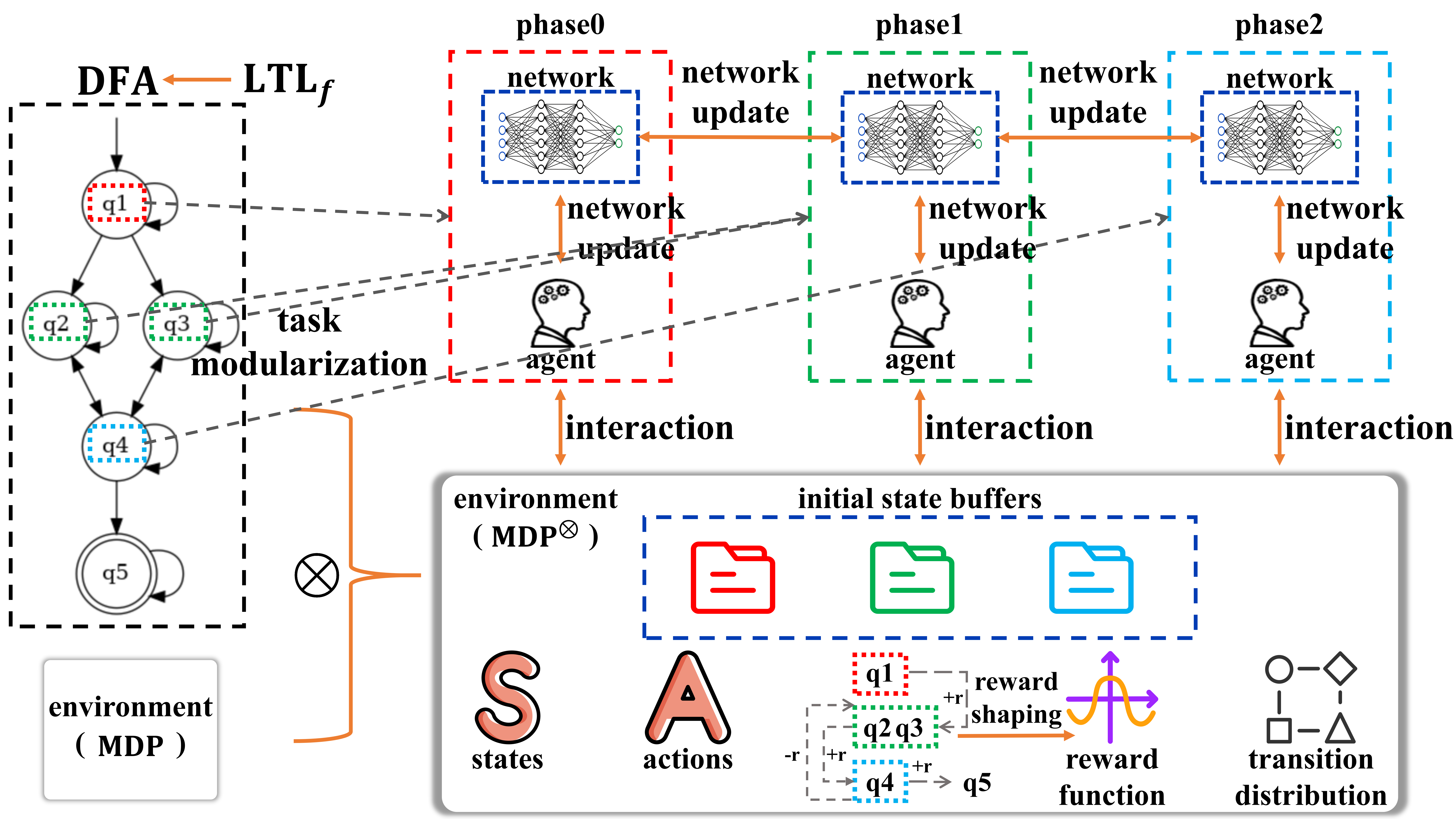}
\caption{Overview of the framework ParMod} 
\label{overview}
\end{figure}

In this section, the \textbf{Par}allel and \textbf{Mod}ular RL framework, referred to as ParMod, is illustrated in detail. The intuition behind our approach is that, by modularizing the NMT into sub-tasks, multiple agents are simultaneously created to learn sub-tasks, with each agent focusing on one sub-task. A comprehensive policy will be synthesised from sub-policies ultimately. Figure \ref{overview} gives an overview of the proposed framework. The NMT specified by an LTL$_f$ formula is transformed to a DFA. The training environment is modeled as a product MDP via synchronising the DFA with the original environment. Modularizing a task essentially means dividing that task into different phases (sub-tasks) based on the classification of DFA states, determined before training. Each task phase corresponds to a category of DFA states. For example, we assume there are three categories of the DFA on the left hand side of Figure \ref{overview}, i.e., $\left \{  q1\right \}, \left \{ q2,q3 \right \}, \left \{ q4 \right \} $, dividing the task into three phases. $q2$ and $q3$ are in the same category since they both need two steps to reach the accepting state $q5$. In ParMod, task phases are trained in parallel. An agent is responsible to handle a task phase by a separate deep network. Totally, the three agents corresponding to the three categories work in collaboration in Figure \ref{overview}.

Most of the task phases are located at intermediate and different positions of a global task. Therefore, a task phase should start from an initial state that may differ from other task phases. To this end, each agent maintains its own initial state buffer (e.g., three initial state buffers at the bottom right in Figure \ref{overview}). It randomly selects an initial state from its buffer at the beginning of an episode and interacts with the environment. The update course for a network is not only influenced by the sampled experiences, but also connected with other networks. In addition, the reward function is reshaped base on the classification result in order to provide appropriate reward signals for each task phase.



\subsection{Modularizing Non-Markovian Tasks}

\begin{figure}[h]
\centering
\includegraphics[width=0.25\textwidth]{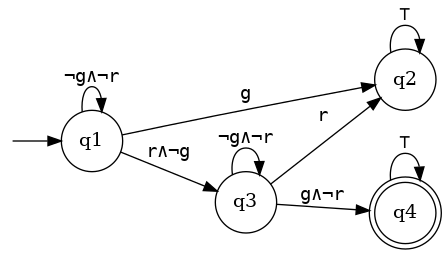}
\caption{DFA of $\varphi = \left ( \neg r \wedge \neg g \right ) \text{U}((r \wedge \neg g) \wedge  \bigcirc ((\neg r \wedge  \neg g)\text{U} (g \wedge \neg r))) $. $\neg g \wedge \neg r, g \wedge \neg r, r \wedge \neg g$ are logical representations of $\{g, r\}, \{ g\}, \{ r \}$ respectively. $g$ represents $\{ g, r \}$ or $\{ g \}$, similar to $r$.}
\label{DFA}
\end{figure}

A modular RL approach is presented in \cite{hasanbeig2020deep} (Modular DDPG). Aiming at synthesising policies of infinite behaviors, the authors construct a similar product MDP with LDBA which results from standard LTL. Each state of LDBA is roughly considered as a ``task divider'', and each transition jumping from one state to another as a ``sub-task''. Taking Figure \ref{DFA} as an example, Modular DDPG modularizes the entire task into four task dividers $q1,q2,q3,q4$, and four sub-tasks, i.e., reaching $g, r, r\wedge \neg g, g\wedge \neg r$, with current DFA state being $q1, q3, q1, q3$ respectively.

However, the efficiency and scalability of Modular DDPG will be severely affected as the size of LDBA increases. Inspired by the idea of Modular DDPG, we design a subtler mechanism for identifying task dividers. DFA states are classified into multiple categories, and each category is viewed as a ``task phase" which is akin to a task divider. 

\begin{algorithm}
\small
\caption{Task Phase Classification (TPC)}
\label{algorhtim: TPC}
\textbf{Input:} 
    $\mathcal{A} =\langle Q,\Sigma,\delta,q_0,F \rangle, N, E=\emptyset$ \\
\textbf{Output:} 
     $rank: Q \rightharpoonup \mathbb{N}$ 
\begin{algorithmic}[1]
\For{$q \in Q \setminus F$}
    \State $d(q) \gets$ compute the average distance from $q$ to $F$
\EndFor
\State $E \gets \{ q \mid d(q) = \infty \text{ and } q \in Q \setminus F \} $
\For{$q \in Q \setminus F \setminus E$}
\State $rank(q) \gets \left\lfloor N-1-\frac{(d(q)-d_{min})(N-1)}{(d_{max}-d_{min} )} + \frac{1}{2}\right\rfloor$
\EndFor
\State \textbf{return} $rank$
\end{algorithmic}
\end{algorithm}

Algorithm \ref{algorhtim: TPC} shows the workflow of Task Phase Classification (TPC). Given a DFA $\mathcal{A}=\langle Q,\Sigma,\delta,q_0,F \rangle$ and the number of categories $1 \le N \le |Q|$, Algorithm 1 returns the ranks of $Q$ ($0 \le rank(q) \le N-1, q \in Q \setminus F \setminus E$) which indicates the degree of task completion as well as categories of $Q$. Let $q \in Q$, $d(q)$ is the average distance of non-looping paths from $q$ to accepting states (line 2). We figure out the error states $E$ which have an infinite distance to accepting states (line 4). Finally, the rank of state $q \in Q \setminus F \setminus E$ is calculated in lines 5-6. Here $d_{max}$ or $d_{min}$ is the maximum or minimum distance to $F$ among $\{d(q)\mid q\in Q \setminus F \setminus E\}$. The interval $d_{max} - d_{min}$ is equally divided into $N$ sub-intervals, corresponding to the ranks $\{0,\ldots,N-1\}$. Note that the ranks help to identify task phases whose scope excludes $E$ and $F$. The reason is that an episode is terminated as long as either an error state or an accepting state is encountered.



\begin{myexample}[rank of DFA states]\label{ex: TPC}
     Consider the DFA in Figure \ref{DFA}, the average distance of $q1$ is $2$, and that of $q3$ is $1$. Hence we know $q3$ has higher degree of task completion over $q1$. However, they may still have the same rank. In other words, the classification of DFA states is not unique. For instance, suppose $N = 1$, in which case $rank(q1) = rank(q3) = 0$. Furthermore, $q2$ and $q4$ are irrelevant to rank.
\end{myexample}

ParMod establishes a shared deep network for each category, which is used to synthesise the sub-policy for the corresponding task phase. What we mean by ``shared" is that every network is allowed to be seen by all agents. The average distance is introduced as a criterion for evaluating the degree of task completion, i.e., how far away from the accepting states through certain DFA state. Moreover, the classification result can also be utilized to reshape the reward.

\subsection{Parallel Training for Modular Task Phases}

Modular DDPG makes use of multiple networks to solve complex NMTs. Unfortunately, it adopts a sequential update mechanism for networks. That is, only one agent explores the state space of the environment and only one network updates at a time. We are of the opinion that this does not help the sample efficiency, and instead, will make training time become the bottleneck to a great extent.

In our framework, accomplishing an NMT is equivalent to facilitate an accepting DFA trace to break into several segments. The framework ensures efficiency, in this way, even with an extremely large number of DFA states, we can still find appropriate number of categories to prevent excessive space occupation and low efficiency issues of networks.

Given a product MDP $M^\otimes =\left \langle S^\otimes,A,R^\otimes,P^\otimes,\gamma,s^\otimes_0\right \rangle$, and the number of categories $N$. In order to improve sample efficiency, ParMod utilizes a group of agents $\mathfrak{A} = \{ \varsigma_0, \ldots, \varsigma_{N-1} \}$ to train task phases in parallel. Each agent is responsible for training a task phase with a network. In particular, agent $\varsigma_i$ is bound with a unique initial state buffer $\mathcal{B}_i$ which is ready to store suitable initial states for $\varsigma_i$. Define all initial state buffers as $\emph{ISB} = \{ \mathcal{B}_0, \ldots, \mathcal{B}_{N-1} \}$. A product state $s^\otimes = \langle s, q \rangle \in S^\otimes$ generated during the training process will have a chance to be put in the corresponding initial state buffer $\mathcal{B}_{rank(q)}$. Agent $\varsigma_i$ randomly samples an initial state from $\mathcal{B}_i$ when resetting the environment.

For preventing an agent's exploration to other task phases (deviating from its own task phase), an additional termination condition is introduced: task phases are different between two adjacent product states of $M^\otimes$, i.e., $P(\langle s_{t+1}, q_{t+1} \rangle \mid \langle s_{t}, q_{t} \rangle, a_t) \neq 0$ and $rank(q_t) \neq rank(q_{t+1})$. Suppose the current product state and the next one are respectively $\langle s_{t}, q_{t} \rangle$ and $\langle s_{t+1}, q_{t+1} \rangle$. If the task phase changes, agent $\varsigma_{rank(q_t)}$ will store $\langle s_{t+1}, q_{t+1} \rangle$ in $\mathcal{B}_{rank(q_{t+1})}$ to initialize the environment for agent $\varsigma_{rank(q_{t+1})}$. As an illustration, the agent inside the red dashed box in Figure \ref{overview} is responsible for training the first task phase. When an episode is terminated and the current product state is of the form $\langle s_t, q2 \rangle$ or $\langle s_t, q3 \rangle$, the agent needs to store $\langle s_t, q2 \rangle$ or $\langle s_t, q3 \rangle$ in the second initial buffer for initializing the second task phase (green dashed box).



Algorithm \ref{algorhtim: ParMod} shows the pseudo code of ParMod. The input comprises a product MDP $M^\otimes$, a DFA $\mathcal{A}$ and the number of categories $N$. Algorithm \ref{algorhtim: ParMod} applies a popular RL training framework, Actor-Critic \cite{DBLP:conf/nips/KondaT99}, which is a mixed approach on both value-based method and policy-based method. Nonetheless, it should be noted that our approach can be easily instantiated by replacing mainstream RL algorhtims e.g., DDPG \cite{DBLP:conf/icml/SilverLHDWR14}, SAC (Soft Actor-Critic) \cite{DBLP:conf/icml/HaarnojaZAL18}, PPO, and even DDQN (Double DQN) \cite{DBLP:conf/aaai/HasseltGS16} that is not a typical Actor-Critic style RL algorithm.

\begin{algorithm}
\caption{ParMod}
\small
\label{algorhtim: ParMod}
\textbf{Input:} 
    $M^\otimes \!=\!\left \langle S^\otimes,A,R^\otimes,P^\otimes,\gamma,s^\otimes_0\right \rangle ,  \mathcal{A} \!=\!\langle Q,\Sigma,\delta,q_0,F \rangle , N$ \\
\textbf{Initialization} 
\begin{algorithmic}[1]
\State $rank \gets$ TPC$(\mathcal{A}, N)$
\State \leftline{initialize $N$ actor and critic networks} $\left\{ \Theta_{\theta_k}(a\mid s^\otimes), \Omega_{\omega_k}(s^\otimes,a) \mid 0 \leq k \leq N-1 \right\}$
\State initialize $N$ initial state buffers $\left\{ \mathcal{B}_k = \emptyset \mid 0 \leq k \leq N-1 \right\}$
\State initialize $N$ experience buffers $\left\{ \mathcal{R}_k = \emptyset \mid 0 \leq k \leq N-1 \right\}$
\State create and execute $N$ training threads $T_0 \parallel T_1 \parallel \cdots \parallel T_{N-1}$
\end{algorithmic}
\textbf{Training Thread} \\
\textbf{Reset($T_k$)} /* reset the environment state in thread $T_k$ */
 \begin{algorithmic}[1]
 \addtocounter{ALG@line}{5}
\If{$k>0$}
\State $s_{init}^\otimes=\left \langle s, q \right \rangle \in \mathcal{B}_{k}$ /* sample from $\mathcal{B}_{k}$ */
\Else
 \State $s_{init}^\otimes = s_{0}^\otimes=\left \langle s_0, q_0 \right \rangle$ or $s_{init}^\otimes=\left \langle s, q \right \rangle \in \mathcal{B}_{0}$ /* reset by original initial state or sample from $\mathcal{B}_{0}$ */
\EndIf
\State \textbf{return} $s_{init}^\otimes$
\end{algorithmic}
\textbf{Execute($T_k$)} /* perform the following operations for thread $T_k$ */
\begin{algorithmic}[1]
\addtocounter{ALG@line}{11}
\State $s_{1}^\otimes=\left \langle s_1, q_1 \right \rangle $ = \textbf{Reset($T_k$)}
\For{$t\gets 1$ \textbf{to} $total\_steps$}
\State choose $a_t\sim \pi_{\theta_{rank(q_t)}}(s_t^\otimes )$
\State take $a_t$, observe $s_{t+1}^\otimes = \langle s_{t+1}, q_{t+1} \rangle$ and receive $r_t$
\State \textbf{if} $q_{t+1} \in Q \setminus F \setminus E$ and $rank(q_{t+1}) \ne k$ \textbf{then} save $s_{t+1}^\otimes$ in  $\mathcal{B}_{rank(q_{t+1})}$
\State reshape reward $r_t\gets r_t + \gamma\rho(q_{t+1}) - \rho(q_t)$
\State save experience $e=\langle s^\otimes_t,a_t,r_t,s^\otimes_{t+1} \rangle $ in $\mathcal{R}_k$
\State \textbf{if} $q_{t+1} \in F\cup E$ or $rank(q_{t+1}) \ne k$ \textbf{then} $s_{t+1}^\otimes=\left \langle s_{t+1}', q_{t+1}' \right \rangle$ = \textbf{Reset($T_k$)}
\If{$t\bmod train\_interval=0$} 
\State sample experience batch $\left \{\langle s^\otimes_i,a_i,r_i,s^\otimes_{i+1} \rangle  \right \}_{i=1,\ldots,n}$ from $\mathcal{R}_k$
\State calculate critic loss$\quad\quad\quad\quad\quad\quad\quad\quad\quad\quad\quad\quad\quad$  $ L_{\Omega_{\omega_k}} = \frac{1}{N} \sum_{i} \Big( \Omega_{\omega_k}(s^\otimes_i, a_i) - (r_i + \gamma  \Omega_{\omega_{k'}}(s^\otimes_{i+1}, a_{i+1})) \Big)^2$ with $s_{i+1}^\otimes=\left \langle s_{i+1}, q_{i+1} \right \rangle, rank(q_{i+1})=k'$
\State calculate actor loss $\quad\quad\quad\quad\quad\quad\quad\quad\quad\quad\quad\quad\quad$ $L_{\Theta_{\theta_k}} = -\frac{1}{N} \sum_{i} \log(\Theta_{\theta _k}(a_i|s^\otimes_i)) \cdot \Omega_{\omega_k}(s^\otimes_i, a_i)$
\State update networks $\Omega_{\omega_k}$ and $\Theta_{\theta_k}$
\EndIf
\EndFor
\end{algorithmic}
\end{algorithm}

To begin with (line 1), the task is modularized by Algorithm \ref{algorhtim: TPC} into $N$ task phases. Line 2 initializes $N$ global actor and critic networks, lines 3-4 initialize $N$ initial state buffers and experience buffers for $N$ training threads, configured as fixed-length queues. All threads are created and executed concurrently in line 5.

The reset operation of a thread is nontrivial. For the thread corresponding to the first task phase, the initial state is obtained by sampling from $\mathcal{B}_0$ with a given probability or providing just the original initial environment state (line 9). For the rest of the threads, the initial state is reset by randomly sampling from the associated initial state buffers (line 7).


\begin{figure}[t]
\centering
\includegraphics[width=0.48\textwidth]{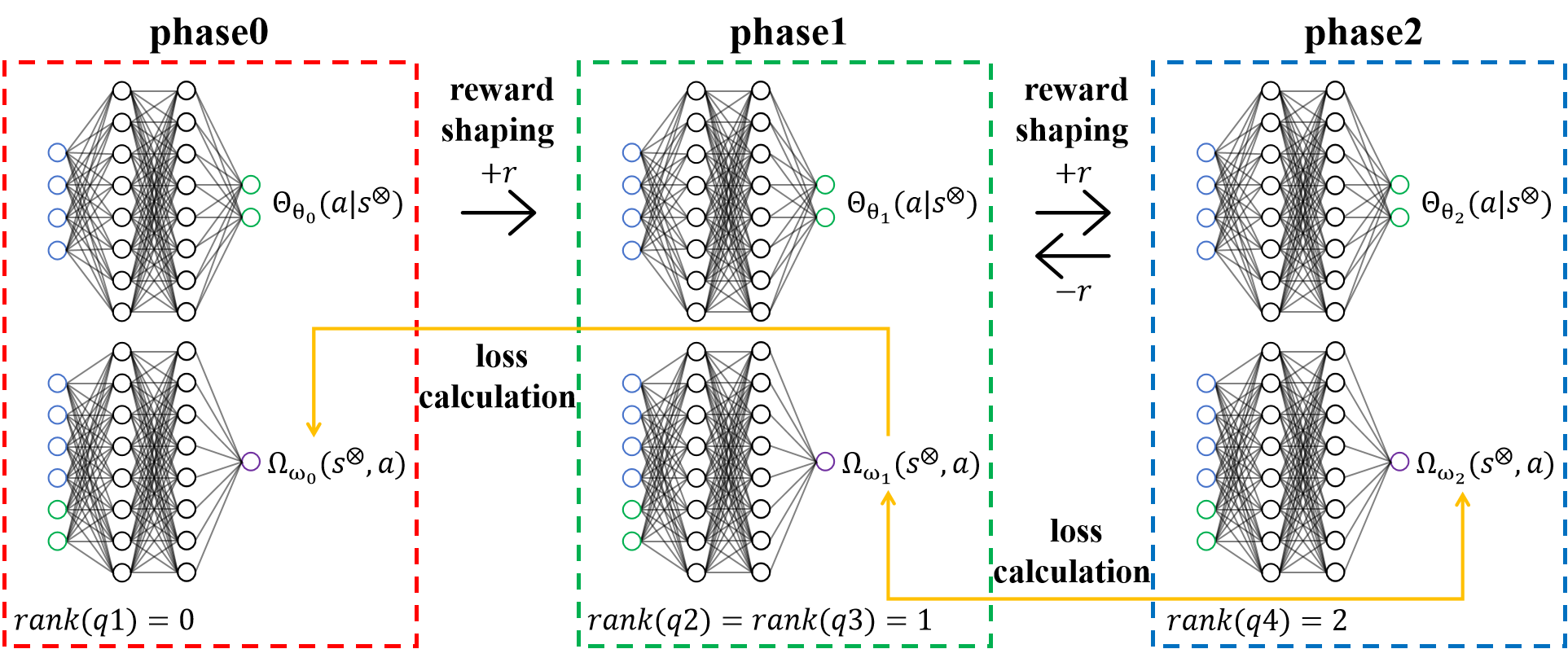}
\caption{Relationship of task phases in ParMod} 
\label{alg2}
\end{figure}

Lines 12-26 describe the learning procedure of a single thread $T_k$. Line 16 checks whether the task phase is changed when interacting with the environment. If so, the next product state $s_{t+1}^\otimes = \langle s_{t+1}, q_{t+1} \rangle$ will be saved in the initial state buffer with the same rank index of $q_{t+1}$, i.e., $\mathcal{B}_{rank(q_{t+1})}$. Line 17 exploits the automated reward shaping technique. Concretely, the reshaped reward function is defined as:
\begin{equation}
R^{\otimes'}(s^\otimes_t, a_t, s^\otimes_{t+1}) = R(s^\otimes_t,a_t,s^\otimes_{t+1}) + \gamma\rho (q_{t+1})-\rho(q_t)\nonumber
\end{equation}
We define the potential function of DFA states $\rho(q)=C/(N-rank(q))$ for $q \in Q \setminus F\setminus E$, where $C$ is a constant. The potentials of the accepting states and the error states are assigned to $C$ and $C/(N+1)$ respectively (the highest and lowest value among $Q$). Line 19 describes the termination conditions for each episode, i.e., the activated DFA state is an accepting state or an error state, or task phase is changed.

Actor and critic networks update in lines 21-24. The loss calculation of critic networks in ParMod is quite different from standard Actor-Critic RL algorithms in essence. The key is to let critic loss rely on rank. Given an experience $\langle s^\otimes_i= \langle s_{i},q_{i} \rangle, a_i,r_i,s^\otimes_{i+1}= \langle s_{i+1},q_{i+1} \rangle \rangle$, the temporal difference error is defined as $\Omega_{\omega_k}(s^\otimes_i, a_i) - (r_i + \gamma  \Omega_{\omega_{k'}}(s^\otimes_{i+1}, a_{i+1}))$. The former critic value is obtained from the critic network with index $k$ being the rank, and the latter critic value is from a critic network with index $k'$ which is not necessary equal to $k$. $k$ and $k'$ are the ranks of $q_{i}$ and $q_{i+1}$ respectively. In other words, we use two possibly different critic networks to make an update for one of them, and a critic network's update may involve another one that is used to assist in training its neighboring task phase. This is due to the fact that $\Omega_{\omega_k'}$ indeed makes a much more accurate evaluation on $s^\otimes_{i+1}$ compared with $\Omega_{\omega_k}$. Figure \ref{alg2} intuitively shows the reward shaping and loss calculation among different task phases, using the DFA in Figure \ref{overview} as an example. A positive or negative reward is received when task phase transfer happens, depending on the rank difference between two task phases.

Finally, the policy is dynamically synthesised by sub-policies, proceeding in the order of product states whose DFA ingredients appear in an accepting 
DFA trace. In the period of validation or test, the agents only need to select actions based on the actor network for the current task phase. By doing this, the additional effort to produce a global strategy network can be avoided.

\section{Correctness and Optimality}
In this section, we discuss the optimality and correctness of policies obtained by ParMod, and provide relevant theoretical proofs.

\begin{mydefinition}

Given an NMRDP $\text{NM}= \langle S, A, R=\langle \varphi, r \rangle, P, \gamma, s_{0} \rangle$ and a DFA $\mathcal{A} =\langle Q,\Sigma,\delta,q_0,F \rangle$ with respect to the reward formula $\varphi$, \textbf{NM is equivalent to the product of $\text{NM}$ and $\mathcal{A}$}, i.e., MDP $M^\otimes =\left \langle S^\otimes,A,R^\otimes,P^\otimes,\gamma,s^\otimes_0\right \rangle$, if there exist two functions $f_1 : S\times Q\longrightarrow S^\otimes$ and $f_2 : S^\otimes \longrightarrow S$ such that

\begin{enumerate}
    \item $\forall s\in S:f_2(f_1(s))=s$.
    \item $\forall s_1,s_2\in S,\forall s_1^\otimes \in S^\otimes$ and $\forall a\in A:$ if $f_2(s_1^\otimes) = s_1$ and $P(s_2\mid s_1, a) >0$, there exits a unique  $s_2^\otimes \in S^\otimes$ such that $f_2(s^\otimes_2)=s_2$ and $ P^\otimes(s^\otimes_2 \mid s^\otimes_1, a) = P(s_2\mid s_1, a)$.
    \item For any feasible trajectory $s_0a_0s_1a_1\ldots s_n$ of $\text{NM}$ and  $s^\otimes_0a_0s^\otimes_1a_1\ldots s^\otimes_n$ of $M^\otimes$, such that $f_1(s_0)=s_0^\otimes$ and $s_i= f_2 (s^\otimes_i)$ for $ 0 \leq i \leq n $, we have $R(s_0\ldots s_{i+1},a_i) = R^\otimes(s_i^\otimes,a_i,s_{i+1}^\otimes)$ for $0\leq i < n$. 
\end{enumerate} 
\end{mydefinition}

\begin{mytheorem}
An NMRDP $\text{NM}= \langle S, A, R=\langle \varphi, r \rangle, P, \gamma, s_{0} \rangle$ is equivalent to the product MDP $M^\otimes =\left \langle S^\otimes,A,R^\otimes,P^\otimes,\gamma,s^\otimes_0\right \rangle$, which is the product of $\text{NM}$ and the DFA $\mathcal{A} =\langle Q,\Sigma,\delta,q_0,F \rangle$ with respect to the reward formula $\varphi$.
\end{mytheorem}
Proof. Recall that every $s^\otimes \in S^\otimes$ has the form $\left \langle s, q \right \rangle$. The reward function of $ \text{\emph{NM}} $ is defined in terms of a pair $ R = \langle \varphi, r \rangle $, where the agent get reward $r$ only when the LTL$_f$ formula is satisfied, i.e.
\begin{equation}
    R(s_0...s_{n+1},a_n)=\begin{Bmatrix}
  r,& \varphi \text{ is satisfied} \\
  0,&\text{ otherwise}
\end{Bmatrix}\nonumber
\end{equation}
 Besides that, the transition probability in $M^\otimes$ is formalized below:

\begin{flushleft}
  $P^\otimes(s_{t+1}^\otimes\mid s_t^\otimes,a_t) = \nonumber$
\end{flushleft}
\begin{equation}
\label{euq1}
  \left\{\begin{matrix}
  P(s_{t+1}\mid s_t,a_t),& s_{t}^\otimes = \langle s_{t}, q_{t} \rangle, s_{t+1}^\otimes = \langle s_{t+1}, q_{t+1} \rangle, \\ &\exists \nu \in \Sigma : \left \langle q_t,\nu ,q_{t+1} \right \rangle \in \delta \text{ and } s_t\models \nu \\
  0,& \text{otherwise}
\end{matrix}\right.
\end{equation}
The reward function in $M^\otimes$ is defined as
\begin{equation}
\label{equ2}
     R^\otimes(s^\otimes_t, a_t, s^\otimes_{t+1})=\left\{\begin{matrix}
  r, & s^\otimes_{t+1}=\left \langle s_{t+1},q_{t+1} \right \rangle, q_{t+1}\in F\\
 0,& \text{otherwise}
\end{matrix}\right.
\end{equation}

First we define $f_1(s) = \left \langle  s,q_0\right \rangle$ and $f_2(s^\otimes) = f_2(\left \langle  s,q\right \rangle)=s$. Then condition 1 is easily verifiable by inspection. For condition 2, let $s_2^\otimes=\left \langle s_2,\delta(q_{1}, s_{1}) \right \rangle$, then the first condition of Equation \ref{euq1} can be satisfied, we have $ P^\otimes(s_{2}^\otimes \mid s_1^\otimes, a) = P(s_2 \mid s_1, a)$ and $f_2(s^\otimes_2)=s_2$. The above proves the existence of $s_2^\otimes$, in the following, we demonstrate the uniqueness. Assume that $s_2^\otimes=\left \langle s^+,\delta(q_{1}, s_{1}) \right \rangle$ where $s^+ \neq s_2 \in S$ and $s^+ \in S$, it cannot hold that $ \forall s_1, s_2 \in S: f_2(s_1^\otimes) = s_1 $ and $ f_2(s_2^\otimes) = s_2 $. Assume that $s_2^\otimes=\left \langle s_2,q^+ \right \rangle$ where $q^+ \neq \delta(q_{1}, s_{1})$, then $P^\otimes(s_{2}^\otimes\mid s_1^\otimes,a) =0 \neq P(s_2\mid s_1,a)$ based on Equation \ref{euq1}. Hence we can conclude the uniqueness.

For condition 3, since $f_1(s_0)=s_0^\otimes$ and $s_i= f_2 (s^\otimes_i)$ for $ 0 \leq i \leq n $, we have $s^\otimes_{i+1} = \left \langle s_{i+1},\delta(q_{i},s_{i}) \right \rangle $ according to condition 2. Then a 
DFA trace $q_0\dots q_n$ can be induced by state trace $s_0\ldots s_n$. We can discuss condition 3 in two cases. When $ \varphi $ is satisfied after the trajectory $ s_0a_0s_1a_1\ldots s_n $, reward $r$ is received in $ \text{\emph{NM}}$ based on $ R = \langle \phi, r \rangle $. Namely, we have the DFA trace $ q_0, q_1, \ldots, q_{n+1}$ such that $ q_{n+1} \in F $. Hence the reward also $r$ in $ M^\otimes $, according to Equation \ref{equ2}. When the task is not completed, the rewards of $ \text{\emph{NM}}$ and $M^\otimes$ are both zero. 

Next, we discuss the equivalence of the optimal policies between $M^\otimes$ and $\emph{NM}$. Let $\pi^\otimes(a\mid s^\otimes) $ be a policy of $M^\otimes$, which determines the probability of an action $a\in A$ at each state $s^\otimes \in S^\otimes$. It is easy to define the equivalent policy of $\emph{NM}$, i.e., $\pi(a\mid s_0\ldots s_n)$. Given any feasible trajectory $s_0a_0s_1a_1\ldots s_n $ of $\emph{NM}$ and DFA trace $q_0\ldots q_{n+1}$ induced by state trace $s_0\ldots s_n$, we have
$\forall a \in A: \pi(a\mid s_0\ldots s_n) = \pi^\otimes(a \mid \left \langle s_n,q_n \right \rangle)$.

\begin{mytheorem}
Given a product MDP $M^\otimes$, let $\pi^{\otimes^{'}}$ be the optimal policy of $M^\otimes$. Then, the policy $\pi^{'}$  that is equivalent to $\pi^{\otimes^{'}}$ is optimal for $\text{NM}$.
\end{mytheorem}
Proof. We prove the following lemma firstly.
\begin{myLemma}
For any feasible trajectory $s_0a_0s_1a_1\ldots s_n$ of $\text{NM}$ and  $s^\otimes_0a_0s^\otimes_1a_1\ldots s^\otimes_n$ of $M^\otimes$, such that $f_1(s_0)=s_0^\otimes$ and $s_i= f_2 (s^\otimes_i)$ for $ 0 \leq i \leq n$, we have

\begin{equation}
\begin{aligned}
Pr_{\pi'}(s_0a_0s_1a_1\ldots s_n) &= Pr_{\pi^{\otimes^{'}}}(s^\otimes_0a_0s^\otimes_1a_1\ldots s^\otimes_n) \\
\mathbb{E} \left[ \sum_{k=0}^{n-1} \gamma ^kR^\otimes(s^\otimes_k,a_k,s^\otimes_{k+1}) \right] &= \mathbb{E} \left[\sum_{k=0}^{n-1} \gamma ^ kR(s_0\ldots s_{k+1},a_k)\right]
\end{aligned}\nonumber
\end{equation}
 where $Pr_\pi(\cdot)$ denotes the probability of generating a trajectory nuder policy $\pi$. 
\end{myLemma}
Proof. $Pr_{\pi'}$ and $Pr_{\pi^{\otimes^{'}}}$ can be transformed in the following form.
\begin{equation}
\begin{aligned}
&Pr_{\pi'}(s_0a_0s_1a_1\ldots s_n)=\pi(a_0\mid s_0)P(s_1\mid s_0,a_0)\pi(a_1\mid s_0s_1)\\&P(s_2\mid s_1,a_1)\ldots\pi(a_{n-1}\mid s_n\ldots s_{n-1})P(s_n\mid s_{n-1},a_{n-1})\\&
Pr_{\pi^{\otimes^{'}}}(s^\otimes_0a_0s^\otimes_1a_1\ldots s^\otimes_n)=\pi^{\otimes^{'}}(a_0\mid s^\otimes_0)P(s_1^\otimes\mid s_0^\otimes,a_0)\pi^{\otimes^{'}}\\&(a_1\mid s_1^\otimes)P(s_2^\otimes\mid s_1^\otimes,a_1)\ldots\pi^{\otimes^{'}}(a_{n-1}\mid s_{n-1}^\otimes)P(s_n^\otimes\mid s^\otimes_{n-1},a_{n-1})
\end{aligned}\nonumber
\end{equation}
 It is easy to obtain that $ P^\otimes(s^\otimes_{i+1} \mid s^\otimes_i, a) = P(s_{i+1}\mid s_i, a)$ and $R(s_0\ldots s_{i+1},a_i) = R^\otimes(s_i^\otimes,a_i,s_{i+1}^\otimes)$ for $0 \leq i < n$ according to Conditions 2 and 3. Then Lemma 1 can be easily verifiable by inspection.

 In the following, we demonstrate Theorem 2 by contradiction. Since $\pi^{\otimes^{'}}$ is the optimal policy of $M^\otimes$, at each state $s^\otimes \in S^\otimes$ we have 
\begin{equation}
\label{equ3}
\pi^{\otimes^{'}}=\operatorname{argmax}_{\pi^{\otimes^{'}}\in D^\otimes}\mathbb{E} \left[\sum_{k=0}^{n-1} \gamma ^ k R^\otimes(s^\otimes_k,a_k,s^\otimes_{k+1})\right]
\end{equation}

where $D^\otimes$ denotes the set of policies over the state space $S^\otimes$, and $s^\otimes_k,a_k,s^\otimes_{k+1},a_{k+1},\ldots$ is a generic path generated by $M^\otimes$ under policy $\pi^\otimes$. Similarly, the optimal policy of $\text{\emph{NM}}$ maximizes the expected accumulated rewards.
Assume that there exists a policy $ \pi^+ \neq \pi^{'} $ and a possible trace $s_0a_0s_1a_1\ldots s_n $ such that 
\begin{equation}
\mathbb{E}_{s_n\sim \pi^+} \left[\sum_{k=n}^{T-1} \gamma ^ kR(s_0\ldots s_{k+1},a_k)\right]>\mathbb{E}_{s_n\sim \pi^{'}}\left[\sum_{k=n}^{T-1} \gamma ^ kR(s_0\ldots s_{k+1},a_k)\right]\nonumber
\end{equation}
According to Lemma 1, for the trajectory $s^\otimes_0a_0s^\otimes_1a_1\ldots s^\otimes_n$ of $M^\otimes$, such that $f_1(s_0)=s_0^\otimes$ and $s_i= f_2 (s^\otimes_i)$ for $ 0 \leq i \leq n$, we have
\begin{equation}
\mathbb{E}_{s^\otimes_n\sim \pi^{\otimes^+}} \left[\sum_{k=n}^{T-1} \gamma ^ kR(s^\otimes_k,a_k,s^\otimes_{k+1})\right]>\mathbb{E}_{s^\otimes_n\sim \pi^{\otimes^{'}}}\left[\sum_{k=n}^{T-1} \gamma ^ kR(s^\otimes_k,a_k,s^\otimes_{k+1})\right]\nonumber
\end{equation}
This is in contrast with optimality of $\pi^{\otimes^*}$ and then Theorem 2 can be proved.

\begin{mytheorem}
Given a product MDP $M^\otimes =\left \langle S^\otimes,A,R^\otimes,P^\otimes,\gamma,s^\otimes_0\right \rangle$ converted by an NMRDP and a DFA with respect to a reward formula $\phi$. Then the optimal policy $\pi^{\otimes^*}$ of $M^\otimes$ minimizes the expected steps to satisfying $\phi$ when $\gamma<1$ or maximizes the probability of satisfying $\varphi$ when $\gamma=1$. 
\end{mytheorem}
Proof. Equation \ref{equ3} has the form below at initial state $s^\otimes_0$
\begin{equation}
\label{equ4}
\begin{aligned}
    \pi^{\otimes^{*}}(s^\otimes_0) & = \operatorname{argmax}_{\pi^{\otimes^{*}}\in D^\otimes}\mathbb{E} \left[\sum_{k=0}^{T-1} \gamma ^ k R^\otimes(s^\otimes_k,a_k,s^\otimes_{k+1})\right] \\
    & = \operatorname{argmax}_{\pi^{\otimes^{*}}\in D^\otimes}\mathbb{E} \left[\gamma ^ L R^\otimes(s^\otimes_L,a_L,s^\otimes_{L+1})\right]\\
    & = \operatorname{argmax}_{\pi^{\otimes^{*}}\in D^\otimes} P(s^\otimes_0) \gamma^{L} r 
\end{aligned}
\end{equation}
where $L$ is the steps to satisfying $\varphi$ and $P(s^\otimes_0)$ indicates the probability of satisfying $\varphi$. Since reward is only given when $\phi$ is satisfied, when $\gamma < 1$, maximizes the expected accumulated rewards for $\pi^{\otimes^*}$ is equivalent to minimizing the steps according to Equation \ref{equ4}. When $\gamma=1$, then $\pi^{\otimes^{*}}(s^\otimes_0)=P(s^\otimes_0)r$ , hence $\pi^{\otimes^*}$ maximizes the probability of satisfying $\phi$. 

Finally, we discuss whether ParMod can converge to the optimal strategy. Due to the strong generalizability, ParMod can set various algorithms as baselines, so the convergence depends on the specifics of the baseline algorithm. Here, we demonstrate this with ParMod and Q-learning as an example. Similar proofs can be referenced when other algorithms are used in ParMod as baselines.
\begin{mytheorem} Given a product MDP $ M^\otimes = \langle S^\otimes, A, R^\otimes, P^\otimes, \gamma, s^\otimes_0 \rangle $ and the DFA $\mathcal{A} =\langle Q,\Sigma,\delta,q_0,F \rangle$, ParMod with tabular Q-Learning converges to an optimal policy as long as every possible state-action pair is visited infinitely often.
\end{mytheorem}
Proof. In ParMod, each modular Q-function $Q_k$ is updated by the rule
\begin{equation}
\begin{aligned}
&Q'_{k}(s_i^\otimes, a_i) = Q_k(s_i^\otimes, a_i) + \alpha_i(s_i^\otimes, a_i) \cdot \\
&\left[ R(s_i^\otimes, a_i, s_{i+1}^\otimes) + \gamma \max_{b \in A} Q_{k'}(s_{i+1}^\otimes, b) - Q_k(s_i^\otimes, a_i) \right]
\end{aligned}\nonumber
\end{equation}
 where $s_i^\otimes=\left \langle s_i, q_i \right \rangle,k = rank(q_i)$ and $k'=rank(q_{i+1})$.

Similarly, Q-learning has the following update format (referring to \cite{melo2001convergence})
\begin{equation}
\begin{aligned}
&Q'(s_i^\otimes, a_i) = Q(s_i^\otimes, a_i) + \alpha_i(s_i^\otimes, a_i)\cdot 
\\&\left[ R(s_i^\otimes, a_i, s_{i+1}^\otimes) + \gamma \max_{b \in A} Q(s_{i+1}^\otimes, b) - Q(s_i^\otimes, a_i) \right] 
\end{aligned}\nonumber
\end{equation}
Note that $Q(s^\otimes,a)$ actually represents the global Q-function, whose value is equal to $Q_k(s^\otimes, a)$ for each state-action pair.  As Q-learning converges to the optimal Q-function, then the optimal Q-value function for $Q_{rank(q)}$ is is equivalent to the optimal Q-function for $Q(\left \langle  s,q\right \rangle, \cdot)$ in $M^\otimes$. Since ParMod with Q-learning selects actions according to modular Q-functions, it can converges to the same optimal policy.

\section{Evaluation}

We implement ParMod as an open source tool built on top of Stable-Baselines3 (SB3) library, which is a set of reliable implementations of RL algorithms by PyTorch \cite{raffin2021stable}. SAC and PPO from SB3 are utilized to perform most evaluations. The hyper-parameters of SAC and PPO are configured by default in SB3. Source code, input files, dataset, and detailed instructions to reproduce our experiments are available in supplementary materials.

\subsection{Experiment Setup}

\begin{table*}
\centering
		\caption{NMT descriptions of \emph{Waterworld} and \emph{Racecar}}\
        \resizebox{0.9\textwidth}{!}{
		\label{tbl1}
        \begin{tabular}{cc|c|c|c}
\toprule
\emph{Waterworld} & \multicolumn{3}{c}{Informal/Formal descriptions} &\#states/\#trans \\
\midrule
\multirow{4}{*}{Task1} & \multicolumn{3}{c}{red strict-then blue strict-then green strict-then red strict-then green then blue} &\multirow{4}{*}{8/20} \\
& \multicolumn{3}{c}{$(\neg r \wedge \neg b \wedge \neg g) \text{U}((r \wedge \neg b \wedge \neg g) \wedge \bigcirc ((\neg r \wedge \neg b \wedge \neg g)\text{U}((b \wedge \neg r \wedge \neg g) \wedge$ }\\
&\multicolumn{3}{c}{ $\bigcirc ((\neg r \wedge \neg b \wedge \neg g) \text{U} ((g \wedge \neg r \wedge \neg b) \wedge \bigcirc ((\neg r \wedge \neg b \wedge \neg g) \text{U}((r \wedge \neg b \wedge \neg g)\wedge $} \\
&\multicolumn{3}{c}{ $\bigcirc((\neg r \wedge \neg b \wedge \neg g)\text{U} ((g \wedge \neg r \wedge \neg b) \wedge \bigcirc((\neg r \wedge \neg b \wedge \neg g) \text{U} (b \wedge \neg r \wedge \neg g)))) ) )) )) )) $} \\
\multirow{4}{*}{Task2} & \multicolumn{3}{c}{(red strict-then blue strict-then green) and (black strict-then white strict-then grey) } &\multirow{4}{*}{17/65} \\
& \multicolumn{3}{c}{$(\neg r \wedge \neg b \wedge \neg g) \text{U}((r \wedge \neg b \wedge \neg g) \wedge \bigcirc ((\neg r \wedge \neg b \wedge \neg g) \text{U}((b \wedge \neg r \wedge \neg g) \wedge $ }\\
& \multicolumn{3}{c}{$\bigcirc ((\neg r \wedge \neg b \wedge \neg g) \text{U}(g \wedge \neg r \wedge \neg b) )) )) \wedge (\neg B \wedge \neg W \wedge \neg G) \text{U}((B \wedge \neg W \wedge \neg G) \wedge $ }\\
& \multicolumn{3}{c}{$\bigcirc ((\neg B \wedge \neg W \wedge \neg G)\text{U}((W \wedge \neg B \wedge \neg G) \wedge \bigcirc ((\neg B \wedge \neg W \wedge \neg G) \text{U}(G \wedge \neg B \wedge \neg W) )) )) $}\\
\multirow{7}{*}{Task3} & \multicolumn{3}{c}{(red strict-then blue strict-then green) and (black strict-then white strict-then grey) } &\multirow{7}{*}{26/106} \\
& \multicolumn{3}{c}{$(\neg r \wedge \neg b \wedge \neg g \wedge\neg p) \text{U}((r \wedge \neg b \wedge \neg g\wedge\neg p) \wedge \bigcirc ((\neg r \wedge \neg b \wedge \neg g\wedge\neg p)$}\\
& \multicolumn{3}{c}{$\text{U}((b \wedge \neg r \wedge \neg g\wedge\neg p) \wedge \bigcirc ((\neg r \wedge \neg b \wedge \neg g\wedge\neg p) \text{U}((g \wedge \neg r \wedge \neg b\wedge\neg p) \wedge $}\\
& \multicolumn{3}{c}{$\bigcirc((\neg r \wedge \neg b \wedge \neg g\wedge\neg p) \text{U}(p \wedge \neg r \wedge \neg b \wedge \neg g) )) )) )) \wedge (\neg B \wedge \neg W \wedge \neg G \wedge\neg Y)$}\\
& \multicolumn{3}{c}{$\text{U}((B \wedge \neg W \wedge \neg G\wedge\neg Y) \wedge  \bigcirc ((\neg B \wedge \neg W \wedge \neg G\wedge\neg Y) \text{U}((W \wedge \neg B \wedge \neg W\wedge\neg G) \wedge $}\\
& \multicolumn{3}{c}{$\bigcirc ((\neg B \wedge \neg W \wedge \neg G\wedge\neg Y) \text{U}((G \wedge \neg B \wedge \neg W\wedge\neg Y) \wedge$}\\
& \multicolumn{3}{c}{$\bigcirc((\neg B \wedge \neg W \wedge \neg G\wedge\neg Y) \text{U}(Y \wedge\neg B \wedge \neg W \wedge \neg G) )) )) ))$}\\
\midrule
\emph{Racecar} & \multicolumn{3}{c}{Informal/Formal descriptions}\\
\midrule
\multirow{3}{*}{Task4} & \multicolumn{3}{c}{reach areas $g_1$ then $g_2$ then $g_3$ then $g_4$ then $g_5$ then $g_6$ or in the reverse sequence} &\multirow{3}{*}{38/161} \\
& \multicolumn{3}{c}{$\Diamond (g_1 \wedge \bigcirc\Diamond(g_2 \wedge \bigcirc\Diamond(g_3 \wedge \bigcirc\Diamond(g_4 \wedge \bigcirc\Diamond(g_5 \wedge  \bigcirc\Diamond g_6))))) \vee$ }\\
& \multicolumn{3}{c}{$\Diamond (g_6 \wedge \bigcirc\Diamond(g_5 \wedge \bigcirc\Diamond(g_4 \wedge \bigcirc\Diamond(g_3 \wedge \bigcirc\Diamond(g_2 \wedge  \bigcirc\Diamond g_1)))))$ }\\
\multirow{3}{*}{Task5} & \multicolumn{3}{c}{reach areas $g_1$ then $g_2$ then $g_3$ then $g_4$ then $g_5$ then $g_6$ then $g_7$ or in the reverse sequence } &\multirow{3}{*}{51/222}\\
& \multicolumn{3}{c}{$\Diamond (g_1 \wedge \bigcirc\Diamond(g_2 \wedge \bigcirc\Diamond(g_3 \wedge \bigcirc\Diamond(g_4 \wedge \bigcirc\Diamond(g_5 \wedge  \bigcirc\Diamond (g_6 \wedge  \bigcirc\Diamond g_7))))))\vee$ }\\
& \multicolumn{3}{c}{$\Diamond (g_7 \wedge \bigcirc\Diamond(g_6 \wedge \bigcirc\Diamond(g_5 \wedge \bigcirc\Diamond(g_4 \wedge \bigcirc\Diamond(g_3 \wedge  \bigcirc\Diamond (g_2 \wedge  \bigcirc\Diamond g_1)))))) $ }\\
\multirow{3}{*}{Task6} & \multicolumn{3}{c}{reach areas $g_1$ then $g_2$ then $g_3$ then $g_4$ then $g_5$ then $g_6$ then $g_7$ then $g_8$ or in the reverse sequence } &\multirow{3}{*}{66/293} \\
& \multicolumn{3}{c}{$\Diamond (g_1 \wedge \bigcirc\Diamond(g_2 \wedge \bigcirc\Diamond(g_3 \wedge \bigcirc\Diamond(g_4 \wedge \bigcirc\Diamond(g_5 \wedge  \bigcirc\Diamond (g_6 \wedge  \bigcirc\Diamond (g_7 \wedge  \bigcirc\Diamond g_8)))))))\vee$ }\\
& \multicolumn{3}{c}{$\Diamond (g_8 \wedge \bigcirc\Diamond(g_7 \wedge \bigcirc\Diamond(g_6 \wedge \bigcirc\Diamond(g_5 \wedge \bigcirc\Diamond(g_4 \wedge  \bigcirc\Diamond (g_3 \wedge  \bigcirc\Diamond (g_2 \wedge  \bigcirc\Diamond g_1))))))) $ }\\
\midrule
\emph{Halfcheetah} & \multicolumn{3}{c}{Informal/Formal descriptions}\\
\midrule
\multirow{2}{*}{Task7} & \multicolumn{3}{c}{reach $D$ strict-then $C$} &\multirow{3}{*}{4/8} \\
& \multicolumn{3}{c}{$(\neg B \wedge \neg D) \text{U} (D \wedge \bigcirc((\neg C \wedge \neg E) \text{U} C)) $ }\\
\multirow{2}{*}{Task8} & \multicolumn{3}{c}{reach $D$ strict-then $C$ strict-then $B$ } &\multirow{3}{*}{5/11}\\
& \multicolumn{3}{c}{$(\neg B \wedge \neg D) \text{U} (D \wedge \bigcirc((\neg C \wedge \neg E) \text{U} (C \wedge \bigcirc((\neg B \wedge D) \text{U} B)))) $ }\\
\multirow{3}{*}{Task9} & \multicolumn{3}{c}{(reach $D$ strict-then $C$ strict-then $B$) or (reach $B$ strict-then $C$ strict-then $D$) } &\multirow{3}{*}{9/25} \\
& \multicolumn{3}{c}{$(\neg B \wedge \neg D) \text{U} (D \wedge \bigcirc((\neg C \wedge \neg E) \text{U} (C \wedge \bigcirc((\neg B \wedge D) \text{U} B))))  \vee$ }\\
& \multicolumn{3}{c}{$(\neg B \wedge \neg D) \text{U} (B \wedge \bigcirc((\neg A \wedge \neg C) \text{U} (C \wedge \bigcirc((\neg B \wedge D) \text{U} D)))) $ }\\
\bottomrule
        \end{tabular}
        }
	\end{table*}
The experiments were conducted on a PC with an Intel Core i7-13700F CPU (24 cores), 32GB DDR5 RAM, running on Windows 10.

Empirical evaluations of three benchmark problems are provided: \emph{Waterworld}, \emph{Racecar} and \emph{Halfcheetah}, whose state and action spaces are continuous. 
\emph{Racecar} is a classical control domain originated from International Planning Competition 2023 (Probabilistic and Reinforcement Learning Track)\footnote{https://ataitler.github.io/IPPC2023/}. A race car is required to reach a target from an initial position. The circular race track is bounded with lines representing the track bounds, hitting one of the bounds is considered a failure and the episode is terminated. The race car is described as a kinematic 2nd order model. The shape of the track is represented by a series of point connected with lines. Referring to the benchmark in \cite{wawrzynski2009cat}, \emph{HalfCheetah} designs a two-dimensional robot comprising 9 links and 8 joints, including two paws. At each time step, the agent chooses how much force to apply to each joint in order to make the robot move forward or backward. The objective is to apply force to the joints, enabling the robot to move between points in a specified sequence.

\begin{figure}[t]
\centering
\includegraphics[width=0.2\textwidth]{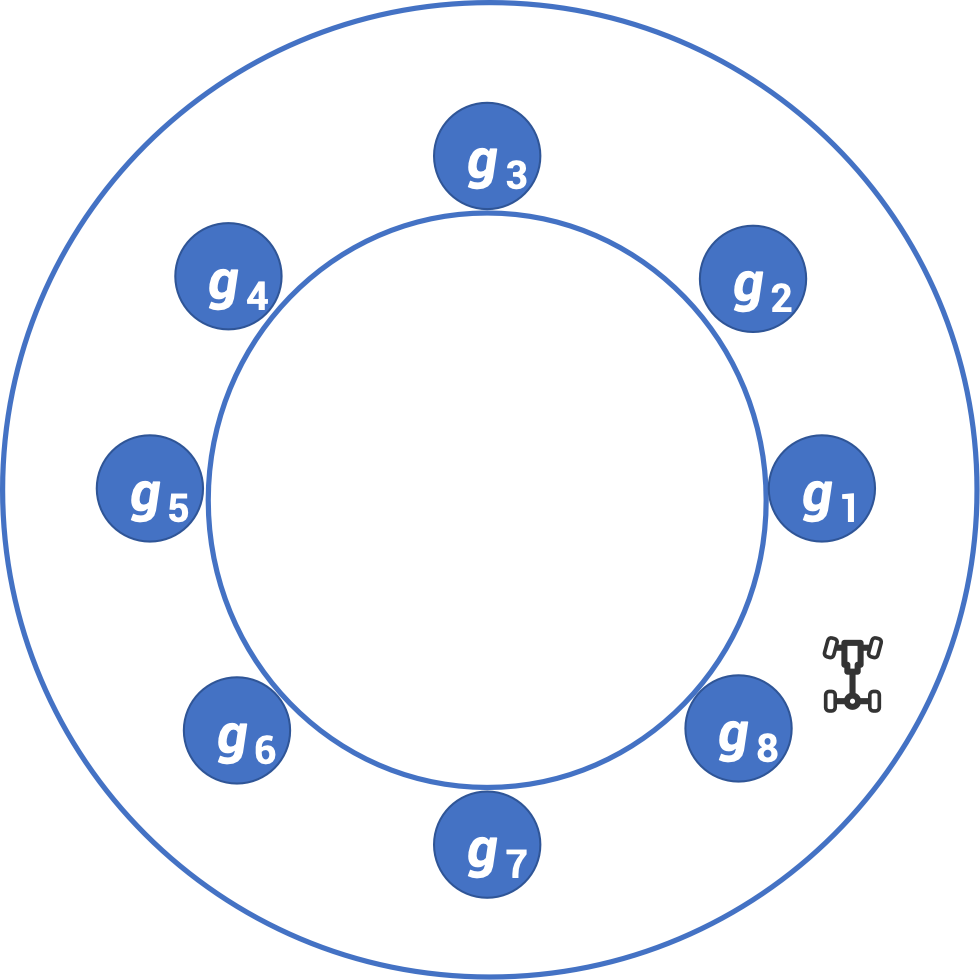}
\caption{Depiction of the \emph{Racecar} problem} 
\label{racecar}
\end{figure}

\begin{figure}[t]
\centering
\includegraphics[width=0.45\textwidth]{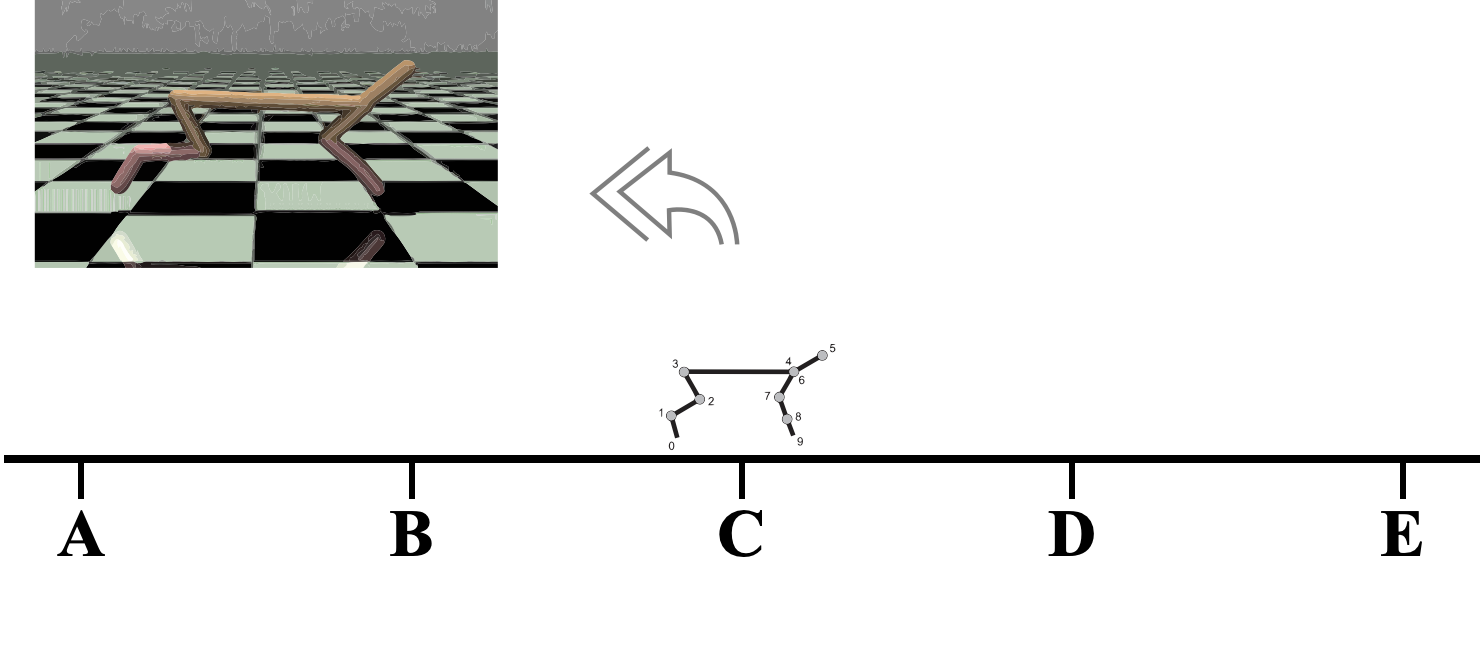}
\caption{Depiction of the \emph{Halfcheetah} problem} 
\label{halfcheetah}
\end{figure}

We randomly generate maps for \emph{Waterworld} with different positions and speeds of balls. The boundary of each map is restricted by $20 \times 20$ or $30 \times 30$, including 3 balls for each color. In Task4-Task6, there exist target areas labeled by $g_{1}$, \ldots, $g_{8}$ along the track for \emph{Racecar}, as shown in Figure \ref{racecar}. The initial position of the cheetah (point $C$) and the locations of the points in the environment are shown in Figure \ref{halfcheetah}. 

The three problems are augmented with several sophisticated NMTs, which are listed in Table 1.  We take Task1, Task4 and Task7 as examples to explain the meanings of the tasks. Task1 requires the agent to touch colored balls in a strict order: red, blue, green, red, green and blue. In Task4, the agent needs to control the car to pass through areas $g_1, g_2, g_3, g_4, g_5, g_6$ in turn or in the reverse order, i.e., $g_6, g_5, g_4, g_3, g_2, g_1$. The major difference between Task1 and Task4 is whether the sub-goals are finished in a strict order or not (strict-then/then). Similar to Task1, the agent in Task7 needs to move from C to D and strict-then C. We exploit the open source tool LTL$_f$2DFA\footnote{http://ltlf2dfa.diag.uniroma1.it/} to map LTL$_f$ formulae into DFAs. The execution time of LTL$_f$2DFA is usually within 0.1s that can be ignored. The last column shows the number of DFA states (\#states) and transitions (\#trans) corresponding to NMTs.

A reward of 100 will be received if a task is completed, regardless of which benchmark is employed. Besides, as is known to all, a policy is evaluated periodically at fixed intervals during training. Since the total reward per episode is either 100 or 0, for better reflecting the optimization trend of a policy, we change the reward function \textbf{only} in the validation process as follows:
\begin{equation}
( \rho(q_{t+1}) - \rho(q_t)) \cdot \left( 1 - (\rho(q_{t+1}) == C/(N+1)) \right) \nonumber
\end{equation}
where $\rho(q_{t+1}) == C/(N+1)$ is a condition used to check whether an error state is activated. In this way, a higher reward indicates a higher rank of the activated DFA state, that is to say a higher degree of task completion.

\subsection{Experimental Results}

In the first experiment, we compare the performance of our approach with other relevant works, i.e., Base, Mod and QRM. Base represents the baseline RL algorithms SAC or PPO. Mod denotes the same modular architecture to Modular DDPG by replacing DDPG with SAC or PPO. 
QRM refers to \emph{Q-learning for reward machines} as discussed in \cite{icarte2022reward}, using SAC as the baseline. Similar to Mod, QRM learns a distinct Q-value function for each DFA state, but it employs counterfactual reasoning to generate synthetic experiences. Note that PPO cannot be used as a baseline for QRM due to the inherent limitations associated with QRM.


We train each task with ParMod, Mod, QRM and Base, repeating 10 times, and collecting the average performance. The number of categories $N$ for Task1-Task3 is set to 6, 10 for Task4-Task6 and 3 for Task7-Task9. Note that the reward shaping technique of ParMod is also applied to Mod, QRM and Base with $C=100$ in the potential function of DFA states.


\begin{figure}[t]
\centering
\includegraphics[width=0.48\textwidth]{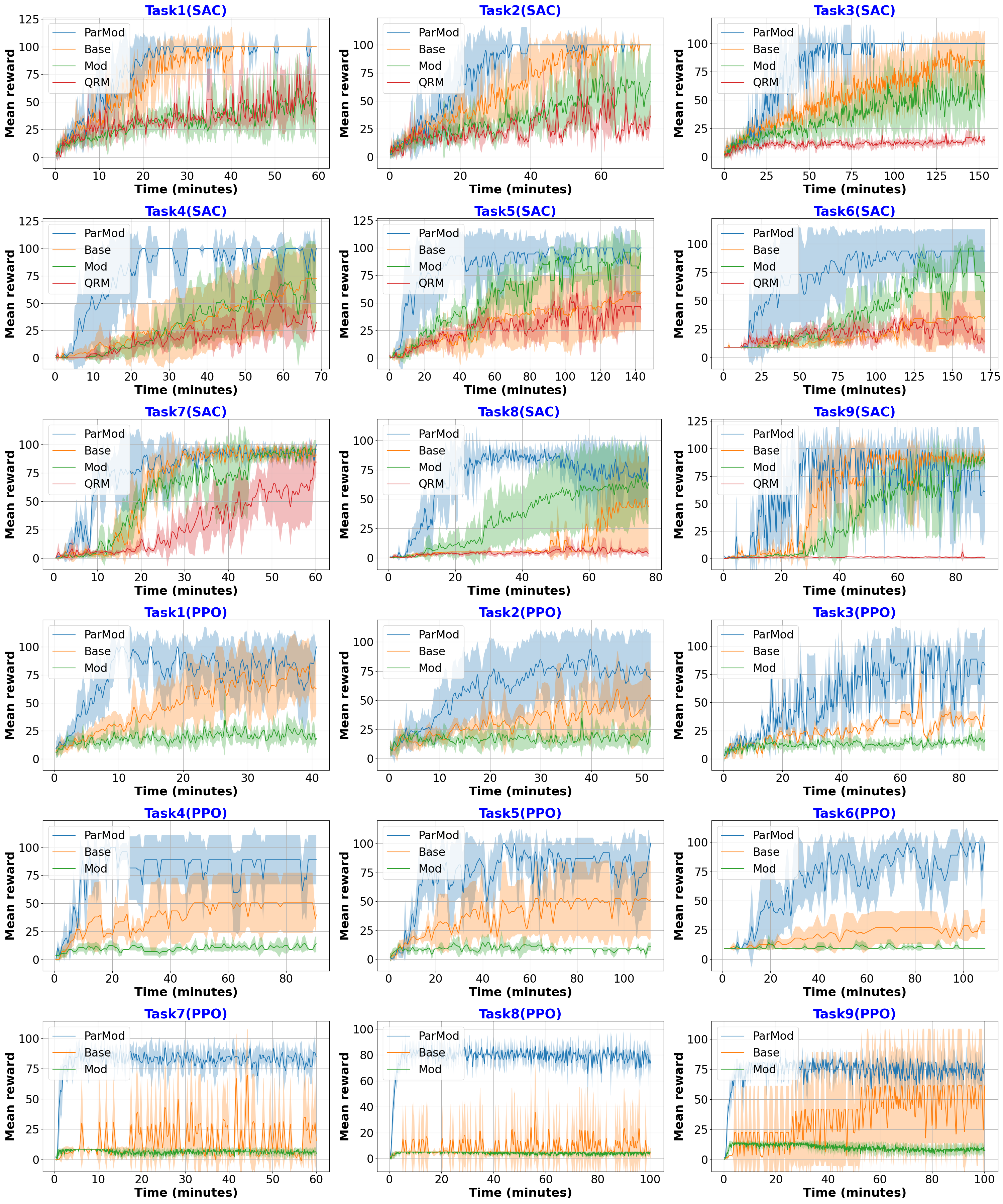}
\caption{Results on the training speed} 
\label{reward}
\end{figure}

Figure \ref{reward} compares ParMod with relative works on the training speed. The horizontal axis represents the training time, and the vertical axis the average rewards per episode. Among all tasks, reward curves of ParMod are the best, substantially surpassing those of QRM, Mod and Base. 

In the case where SAC is the baseline, all the reward curves of ParMod are quickly converged to or near the highest reward. In contrast, relative works cannot converge within the time limit for most tasks, even though their curves tend to rise slowly. It is obvious that the three algorithms may require more time to learn the policy, or they may even fail to do so. The curves of ParMod and Base for Task1 are basically identical. The reason is that Task1 is relatively simple for SAC itself. Moreover, Base exhibits superior performance to Mod except Task5, Task6 and Task 8. We can infer that training task phases without parallelization does not necessarily help, on the contrary, it lowers the efficiency in many cases. Furthermore, QRM exhibites the poorest performance across all tasks. We hypothesize that, within such reward setting, the counterfactual reasoning mechanism produces redundant experiences and reward signals, which consequently diminishes training performance.


When using PPO as the baseline, the results are slightly inferior to those based on SAC in general. This mainly attributes to the difference between the two baselines. Nevertheless, ParMod still shows significantly advantage over other methods. The remarkable \textbf{generalizability} of ParMod can be demonstrated, no matter the underlying RL algorithm is off-policy (SAC) or on-policy (PPO). Mod performs extremely poor on all tasks, with all curves lower than Base. We speculate that this may due to the limited generalizability of Mod applied to on-policy algorithms. 

\begin{table}[h]
    \centering
    \caption{Comparison in success rate, policy quality and convergence rate}
    \resizebox{0.5\textwidth}{!}{%
    \begin{tabular}{ccccccccccc}
    \toprule
        \multicolumn{2}{c}{SAC\textbackslash PPO}&Task1& Task2 & Task3 & Task4& Task5 & Task6 & Task7 & Task8 & Task9  \\
    \midrule
     \multicolumn{2}{c}{Percentage (\%)} & &  & &  &  &  \\
    \midrule
         \multirow{3}{*}{Success rate}&ParMod & \textbf{100\textbackslash100} & \textbf{100\textbackslash100} & \textbf{100\textbackslash100} & \textbf{100\textbackslash100} & \textbf{100\textbackslash100} & \textbf{100\textbackslash100} & 
         \textbf{100\textbackslash100} & 
         \textbf{100\textbackslash100} & 
         \textbf{100\textbackslash100}  \\
         &Base & \textbf{100}\textbackslash 70 & \textbf{100}\textbackslash80 & \textbf{100}\textbackslash40 & 50\textbackslash20 & 40\textbackslash20 & 20\textbackslash0 &
         100\textbackslash30 &
         50\textbackslash30 &
         100\textbackslash60 
         \\
         &Mod & \textbf{100}\textbackslash40 & \textbf{100}\textbackslash20 & \textbf{100}\textbackslash20 & 70\textbackslash0 & \textbf{100}\textbackslash0 & 80\textbackslash0 &
         \textbf{100}\textbackslash0 &
         70\textbackslash0 &
         \textbf{100}\textbackslash0 
         \\
         &QRM & \textbf{100}\textbackslash--- & \textbf{100}\textbackslash--- & \textbf{0}\textbackslash--- & 40\textbackslash--- & 20\textbackslash--- & 0\textbackslash--- &
         80\textbackslash--- &
         0\textbackslash--- &
         0\textbackslash--- 
         \\
    \midrule
     \multicolumn{2}{c}{Length} & &  & &  &  &  \\
    \midrule
         \multirow{4}{*}{Policy quality}&ParMod & \textbf{46.9\textbackslash59.4} & \textbf{52.6\textbackslash64.3} & \textbf{67.7\textbackslash95.8} & \textbf{61.1\textbackslash54.8} & \textbf{71.0\textbackslash62.8} & \textbf{70.6\textbackslash67.0} & 
         \textbf{73.3\textbackslash128.6} &
         \textbf{129.9\textbackslash171.0} &
         \textbf{91.4\textbackslash145.0}\\
         &Base & 55.4\textbackslash69.2 & 61.3\textbackslash69.5 & 109.2\textbackslash112.0 & 77.0\textbackslash59.0 & 80.2\textbackslash101.0 & 97.4\textbackslash--- & 77.7\textbackslash410.5 &
         147.9\textbackslash516.6 &91.5\textbackslash206.5 \\
         &Mod & 54.7\textbackslash79.3 & 57.9\textbackslash139.0 & 75.0\textbackslash149.0 & 70.5\textbackslash--- & 75.4\textbackslash--- & 82.0\textbackslash--- &80.5\textbackslash--- &115.5\textbackslash--- &114.9\textbackslash--- \\
         &QRM & 63.5\textbackslash--- & 71.6\textbackslash--- & ---\textbackslash--- & 123.5\textbackslash--- & 90.0\textbackslash--- & ---\textbackslash--- & 152.3\textbackslash--- & ---\textbackslash--- & ---\textbackslash--- \\
    \midrule
        \multicolumn{2}{c}{Training time (min)} & &  & &  &  &  \\
    \midrule
        \multirow{4}{*}{Conv. rate}&ParMod & \textbf{10.2\textbackslash4.9} & \textbf{12.3\textbackslash9.0} & \textbf{20.6\textbackslash31.4} & \textbf{12.6\textbackslash9.0} & \textbf{21.1\textbackslash17.8} & \textbf{35.8\textbackslash26.1} &
        \textbf{13.8\textbackslash4.5} &
        \textbf{26.6\textbackslash28.7} &
        \textbf{11.1\textbackslash27.9} \\
        &Base & 12.8\textbackslash21.7 & 28.4\textbackslash40.8 & 61.8\textbackslash108.5 & 53.0\textbackslash13.9 & 98.9\textbackslash39.4 & 172.8\textbackslash--- &
        24.6\textbackslash20 &
        71.0\textbackslash32.6 &
        34.3\textbackslash36.1 \\
         &Mod & 28.7\textbackslash35.3 & 28.1\textbackslash38.1 & 76.0\textbackslash90.9 & 54.4\textbackslash--- & 73.1\textbackslash--- & 135.1\textbackslash--- &
         33.2\textbackslash--- &
         58.1\textbackslash--- &
         67.6\textbackslash--- \\
         &QRM & 16.2\textbackslash--- & 51.6\textbackslash--- & ---\textbackslash90.9 & 68.9\textbackslash--- & 130.1\textbackslash--- & ---\textbackslash--- &
         52.4\textbackslash--- &
         ---\textbackslash--- &
         ---\textbackslash--- \\
    \bottomrule
    \end{tabular}
    }
     \label{policy quality}
\end{table}

In order to provide more comprehensive evaluations of our approach, we compare different methods on the metrics of success rate, policy quality and convergence rate. Table \ref{policy quality} reports the statistics with regard to these metrics. The policy quality is measured by the length of a policy (the shorter the better), and the convergence rate is represented by the first time to synthesise a policy during training.

\begin{figure}
\centering
\includegraphics[width=0.48\textwidth]{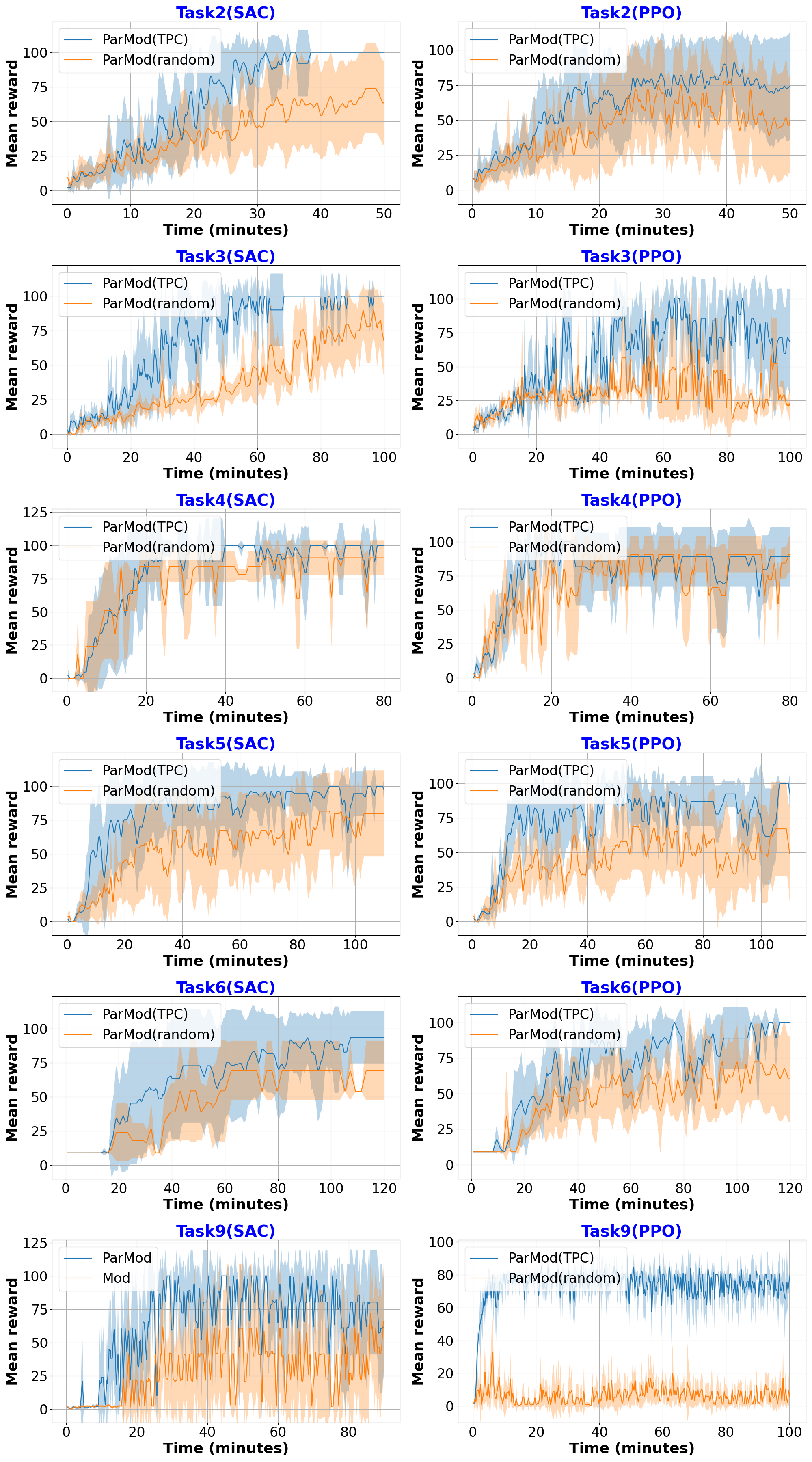}
\caption{Comparison between TPC and random task phase classification} 
\label{random}
\end{figure}

We take 200 runs applying the best learned models, and analyse the success rate for all aforementioned tasks. It is evident that ParMod achieves the perfect success rate of 100\% in all tasks. ParMod is also the winner in policy quality and convergence rate. Here ``---" denotes no success policy has been found. It is worth to mention that, the more complex the task, the more effective for ParMod. For Task1, ParMod with SAC converges 187\% faster on average than the others (122\% over Base, 281\% over Mod and 158\% for QRM). For the more complex Task3, the average convergence speed increases to 334\%, with improvements of 368\% over Mod and 300\% over Base, while QRM fails to find any successful policy. Interestingly, ParMod also shows the strength in policy quality. The reason is that the modular scheme is capable of generating more sub-policies, thereby improving the quality of the global policy. This is why the policy quality of ParMod is much better than Base. There are more policy iterations for each task phase in ParMod, hence the policy quality is also better than Mod and QRM.


The second experiment aims to demonstrate the effectiveness of the proposed classification method TPC. Specifically, to answer the question: whether employing TPC to modularize task phases is more effective than a random classification strategy (referred to as ``random"). Note that random classification of DFA states amounts to associate DFA states with random rank values. We uniformly use the reshaped reward function calculated by TPC. Moreover, only DFA states are randomly classified, while $N$ remains unchanged.
Figure \ref{random} presents a comparison of training speed across most tasks, with the exception of Task1, Task7 and Task8. The reason is that each category in the three tasks contains only one DFA state, hence the classification results for TPC and the random classification strategy are identical. 
The results indicate that TPC indeed helps improve the performance of ParMod (all curves of TPC are higher than ``random"). The main reason is attributed to the scheduling of task phases. The difficulty of learning between any two adjacent task phases may vary a lot provided task phases are arbitrarily divided. TPC is able to reasonably divide task phases based on the degree of task completion, enabling ParMod to construct the potential function for reward shaping in a relatively balanced manner.

\begin{figure}[t]
\centering
\includegraphics[width=0.48\textwidth]{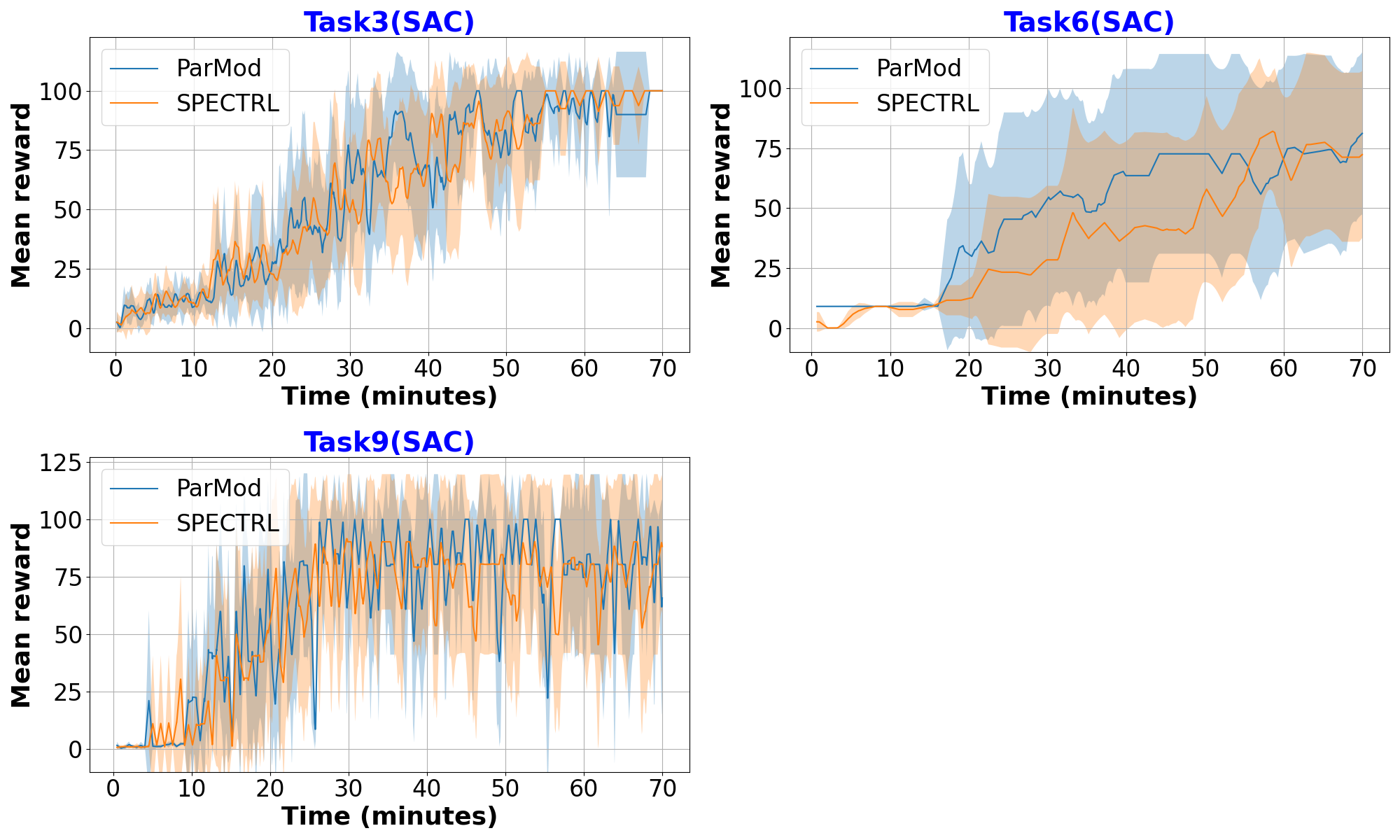}
\caption{Comparison in the training speed: ParMod vs. SPECTRL} 
\label{rs}
\end{figure}

For the reward shaping mechanism in ParMod, which is implemented based on the automaton structure, we also conduct an experiment to compare it with the related reward shaping mechanism in SPECTRL \cite{jothimurugan2019composable}. In contrast to ParMod, SPECTRL not only considers the "distance" factor between automaton nodes but also incorporates the "completeness" of each atomic predicate through quantitative semantics. Figure \ref{rs} illustrates the comparison between our reward shaping mechanism and that of SPECTRL in the three most challenging tasks (Tasks 3, 6, and 9). Note that both methods utilize the parallel module framework in ParMod, differing only in the reward settings. In summary, there is no significant difference between the two methods, and our method is even slightly better than SPECTRL in the three tasks. The outcomes indicate that extending the quantitative semantics of temporal logic to the reward reshaping in ParMod did not improve training speed. Instead, it reduced training performance due to the computation of "completeness" for atomic predicates.

\begin{figure}[t]
\centering
\includegraphics[width=0.48\textwidth]{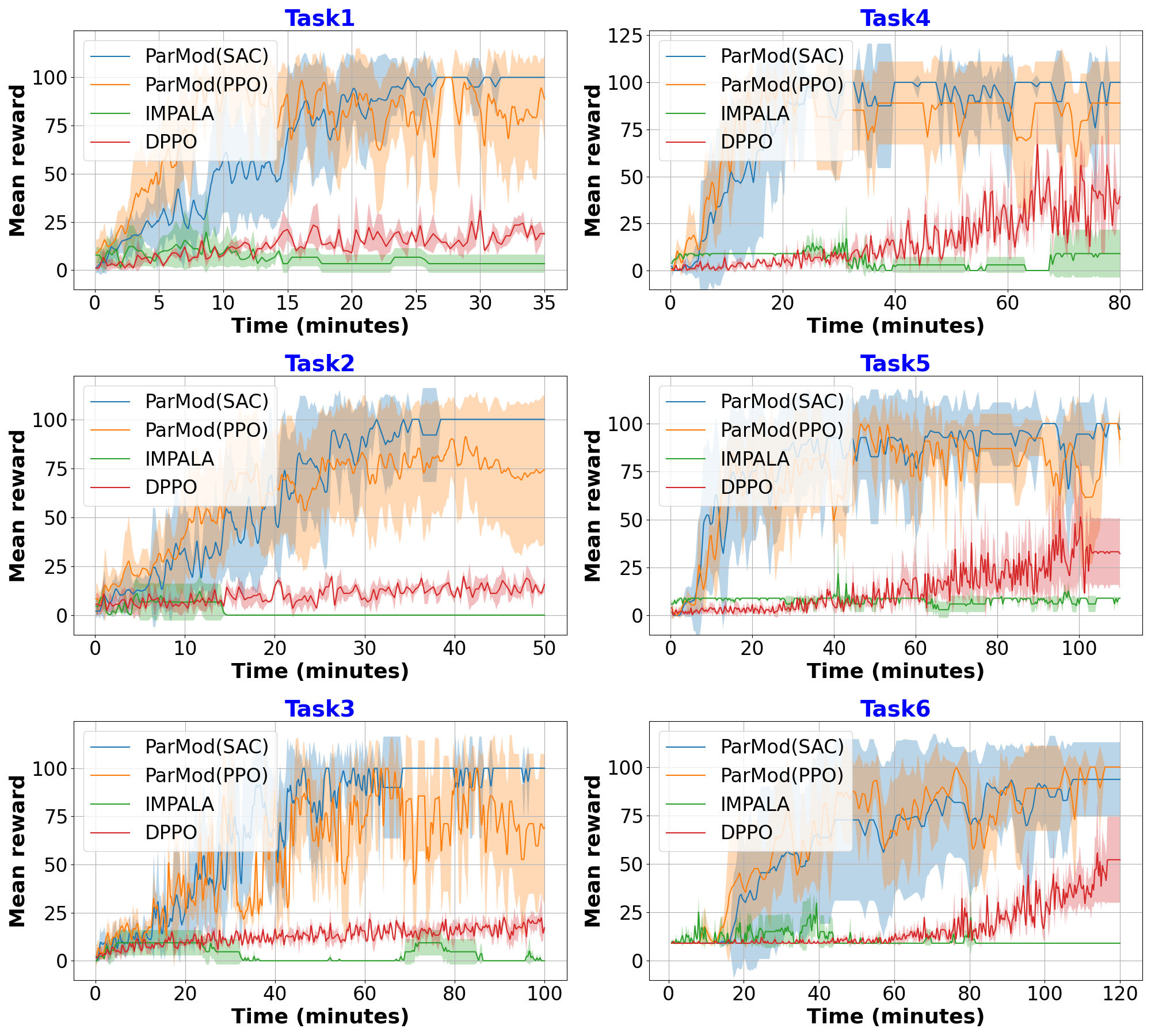}
\caption{Results on the training speed: ParMod (24 CPU cores) vs. SOTA algorithms (150 CPU cores) } 
\label{sota}
\end{figure}

Another experiment is conducted for comparing ParMod with the state-of-the-art distributed RL approaches IMPALA and DPPO. To be fair, we directly run the source code\footnote{https://github.com/ray-project/ray/tree/master/rllib} of IMPALA and DPPO on a distributed architecture equipped with 150 CPU cores, additionally with ParMod's reshaped reward function. Figure \ref{sota} shows the results for all tasks. There is no doubt that IMPALA and DPPO exhibit unsatisfactory performance. They fail to generate policies for all tasks, and the degree of task completion for the learned policies is far insufficient. It can be inferred that, even if distributed RL algorithms can sample experiences in parallel through multiple workers, later task phases are seldom reached when facing with complex NMTs, resulting in inadequate exploration of them. On the contrary, ParMod continuously allocates initial states to later task phases, whose scope of exploration is expanded and frequency of policy iteration is increased. Additionally, each task phase in ParMod is equipped with a separate deep network, which makes learning be targeted to intermediate goals, hence improve the overall training performance. In a word, the powerful performance of ParMod originates from the integration of the parallel learning and automaton-based reward shaping. The cooperation of the two enables to receive appropriate reward signals at each task phase as soon as possible, and either of which is indispensable.

\begin{figure}[t]
\centering
\includegraphics[width=0.4\textwidth]{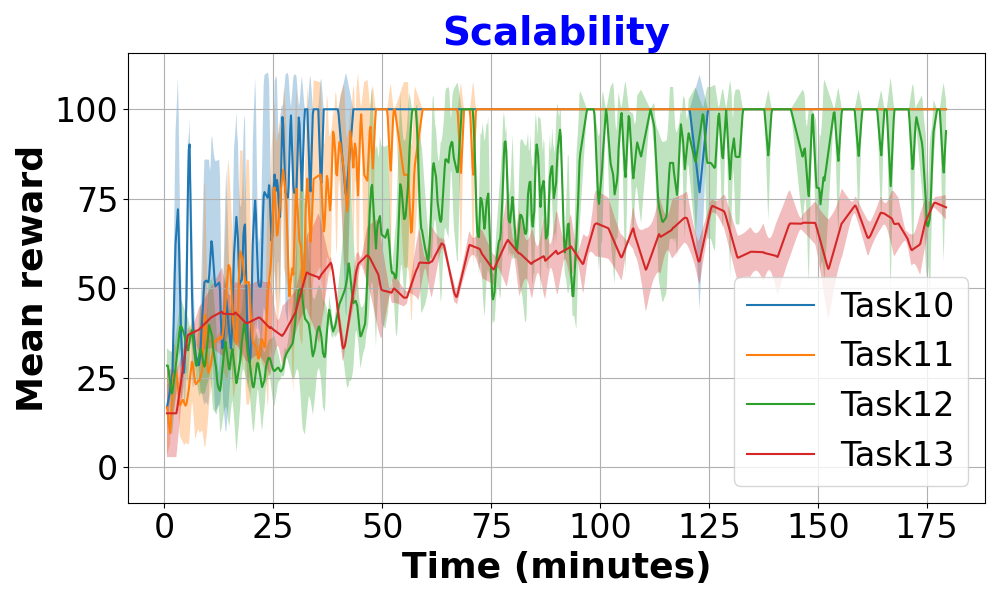}
\caption{Results on the training speed for Task10-Task13} 
\label{scalability}
\end{figure}

The fifth experiment focuses on evaluating the scalability of our approach. We design four additional tasks (Task10-13) of \emph{Waterworld} that the corresponding DFAs are much more complex than Task1-3. The number of DFA states and transitions are at most up to 290 and 1091. Table \ref{table2} provides the detailed information about the four tasks, where $c_1, c_2, \ldots $ represent touching balls of different colors. The scalability of our approach is intuitively shown via the performance of the four tasks in Figure \ref{scalability}. In a word, our approach exhibits high scalability as the complexity of the tasks increases. The reward curves show a trend of gradual decline on learning capability from Task10-13, rather than a dramatically decline. The reward curves of Task 10 and Task11 converge within one hour. Despite the fact that the curves of Task12 and Task13 do not converge within three hours (Task12 near optimality), the continued upward trend implies that they just need more training.

\begin{table}
    \centering
		\caption{NMT Descriptions for Task10-Task12}\
        \resizebox{0.48\textwidth}{!}{
		\label{tbl1}
       \begin{tabular}{cccc}
        \toprule
\emph{Waterworld} & Informal/Formal descriptions &\#states &\#trans\\
\midrule
    \multirow{3}{*}{Task10} & $c_1$ then $c_2$ then $c_3$ ... then $c_8$ or in the reverse sequence & \multirow{3}{*}{65} & \multirow{3}{*}{227}\\
       & $\Diamond (c_1 \wedge \bigcirc\Diamond(c_2 \wedge \bigcirc\Diamond(c_3 \wedge \bigcirc\Diamond(... \wedge\bigcirc\Diamond(c_8)))))\vee$\\
       & $\Diamond (c_8 \wedge \bigcirc\Diamond(c_7 \wedge \bigcirc\Diamond(c_6 \wedge \bigcirc\Diamond(... \wedge\bigcirc\Diamond(c_1)))))$\\
      \multirow{3}{*}{Task11} & $c_1$ then $c_2$ then $c_3$ ... then $c_{11}$ or in the reverse sequence  & \multirow{3}{*}{122}& \multirow{3}{*}{443}\\
      & $\Diamond (c_1 \wedge \bigcirc\Diamond(c_2 \wedge \bigcirc\Diamond(c_3 \wedge \bigcirc\Diamond(... \wedge\bigcirc\Diamond(c_{12})))))\vee$\\
      & $\Diamond (c_{12} \wedge \bigcirc\Diamond(c_{11} \wedge \bigcirc\Diamond(c_{10} \wedge \bigcirc\Diamond(... \wedge\bigcirc\Diamond(c_1)))))$\\
      \multirow{3}{*}{Task12} & $c_1$ then $c_2$ then $c_3$ ... then $c_{14}$ or in the reverse sequence& \multirow{3}{*}{197}& \multirow{3}{*}{731}\\
      & $\Diamond (c_1 \wedge \bigcirc\Diamond(c_2 \wedge \bigcirc\Diamond(c_3 \wedge \bigcirc\Diamond(... \wedge\bigcirc\Diamond(c_{14})))))\vee$\\
      & $\Diamond (c_{14} \wedge \bigcirc\Diamond(c_{13} \wedge \bigcirc\Diamond(c_{12} \wedge \bigcirc\Diamond(... \wedge\bigcirc\Diamond(c_1)))))$\\
      \multirow{3}{*}{Task13} & $c_1$ then $c_2$ then $c_3$ ... then $c_{17}$ or in the reverse sequence& \multirow{3}{*}{290}& \multirow{3}{*}{1091}\\
      & $\Diamond (c_1 \wedge \bigcirc\Diamond(c_2 \wedge \bigcirc\Diamond(c_3 \wedge \bigcirc\Diamond(... \wedge\bigcirc\Diamond(c_{17}))))) \vee $\\
      & $\Diamond (c_{17} \wedge \bigcirc\Diamond(c_{16} \wedge \bigcirc\Diamond(c_{15} \wedge \bigcirc\Diamond(... \wedge\bigcirc\Diamond(c_{1})))))$\\
\bottomrule
        \end{tabular}
        }
        \label{table2}
	\end{table}

\begin{figure}[t]
\centering
\includegraphics[width=0.35\textwidth]{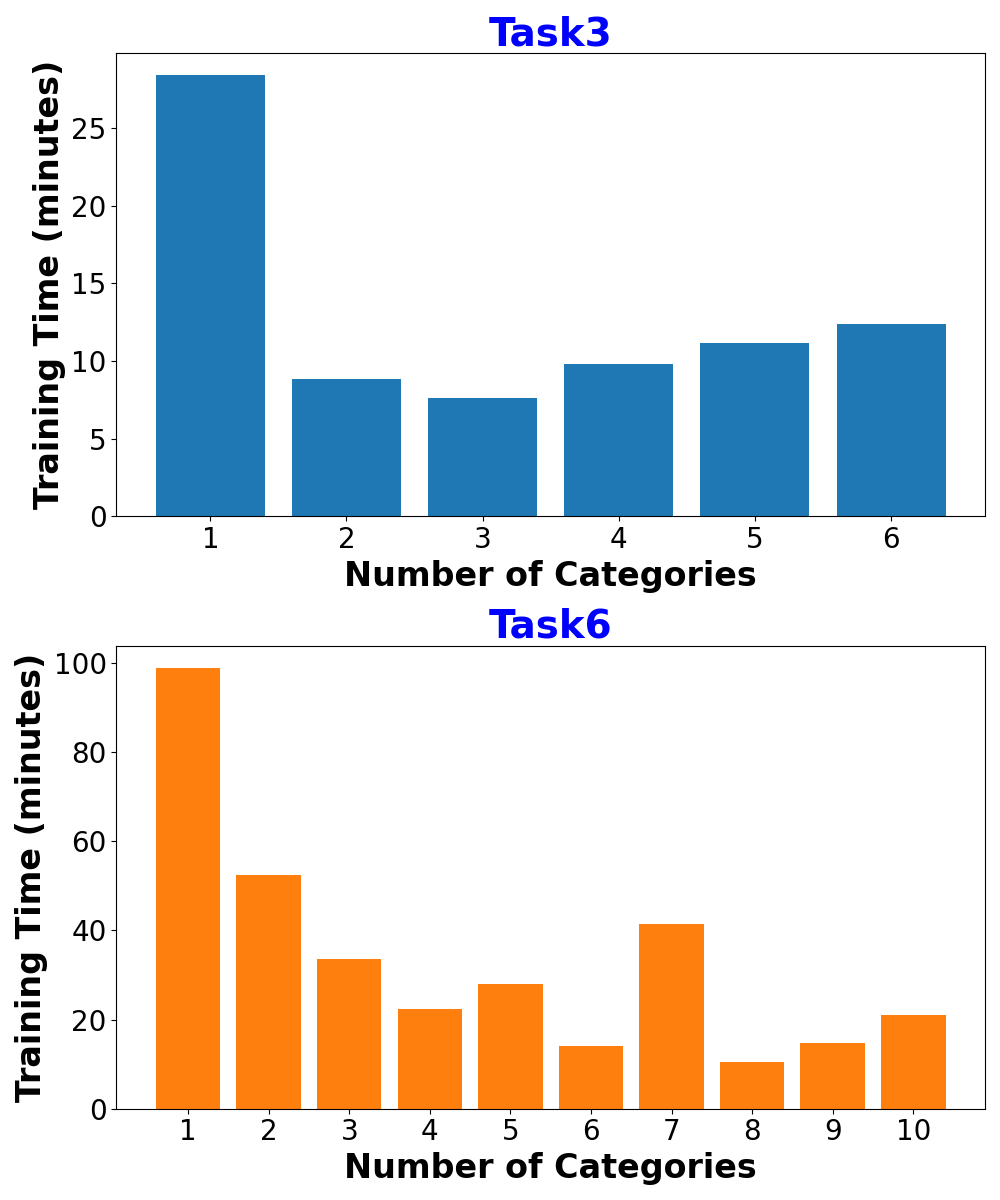}
\caption{The impact of the number of categories $N$ on training speed} 
\label{categories}
\end{figure}

 The next experiment aims to answer the question: how many task phases are the most effective in ParMod. The convergence rate (the first time to synthesise a policy during training) for different number of categories $N$ on Task3 and Task6 is shown in Figure \ref{categories}. It is important to note that the impact of different reward functions on performance is not considered here, hence a consistent reward function is applied in all settings. We find that $N=3$ achieves the best performance for Task3, and $N=8$ is the best in Task6. The training speed of $N=1$ (the lower bound of $N$ in ParMod) is the slowest for both tasks. It means not dividing the NMT, which is equivalent to using the baseline RL algorithm. Conversely, refining an NMT into too many categories is also not a good choice. Since an excessive categories necessitates more networks, thereby increasing the difficulty of coordination among sub-policies. In summary, the optimal number of categories cannot be theoretically determined, which depends on specific tasks and may require parameter tuning.

Finally, considering that most parallel algorithms can improve computational efficiency through distributed systems, we implement a distributed version of ParMod to support parallel learning for a large number of sub-tasks. We exploit \cite{moritz2018ray} as the underlying architecture, which enables seamless scaling of the same code from a single-machine to a cluster. For demonstrating the effectiveness of the distributed ParMod, we compare it with standalone ParMod on the training speed. The distributed ParMod is equipped with 64 CPU cores, and the training parameters are consistent with those of the standalone ParMod. Note that we only implement a basic and plain distributed version of ParMod at this stage. Each agent is assigned 6 CPU cores for learning sub-tasks, while an additional 4 CPU cores are allocated for parameter transmission\footnote{64 CPU cores are enough for Task6, and 40 CUP cores for Task3, since Task3 and Task6 are divided into 6 and 10 task phases respectively.}. In future work, we will integrate state-of-the-art distributed RL algorithms, allowing multiple actors to collect experiences in parallel, thereby supporting ParMod to make the best use of computational resources.

\begin{figure}[t]
\centering
\includegraphics[width=0.48\textwidth]{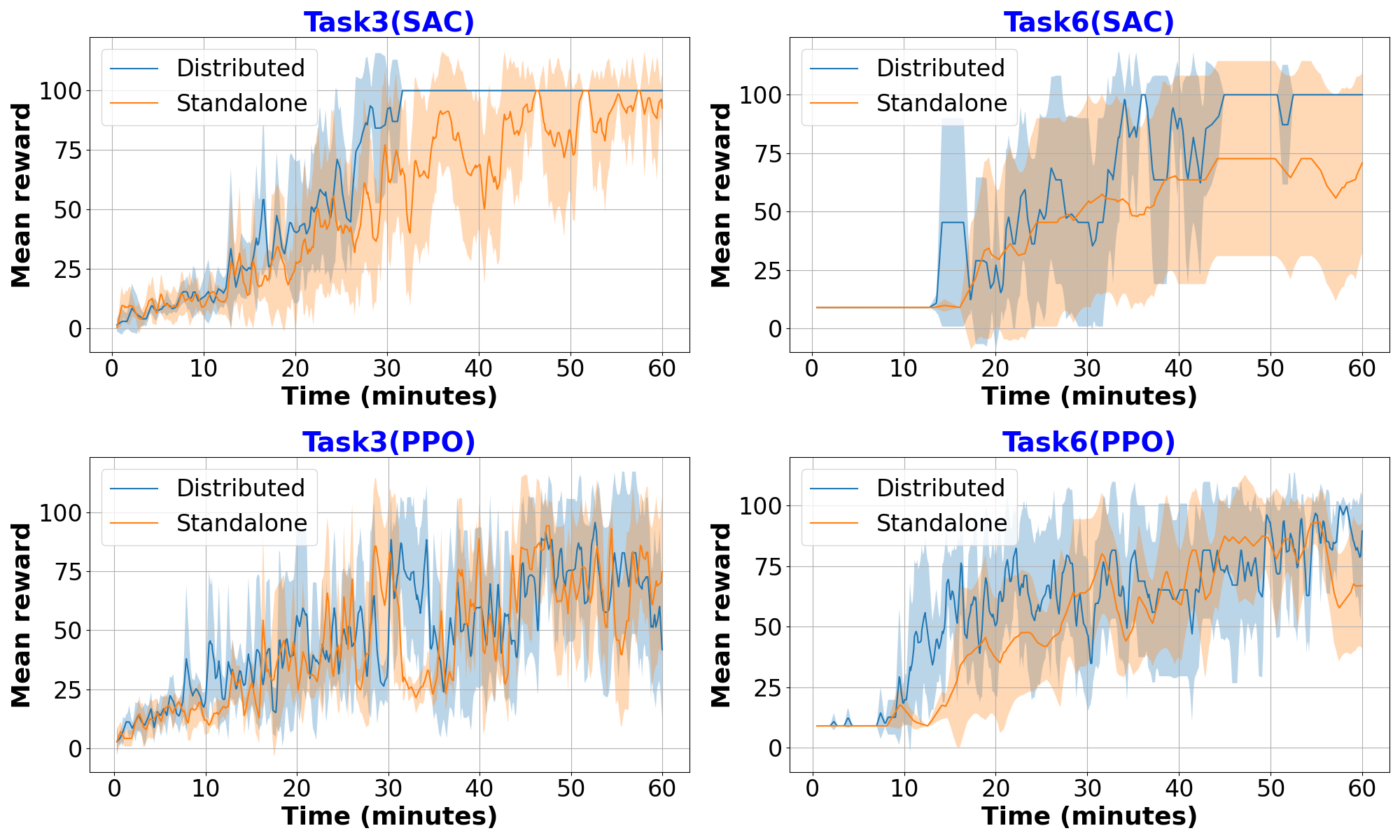}
\caption{Results on the training speed of ParMod: Distributed (64 CPU cores) vs. Standalone (24 CPU cores)} 
\label{fig1}
\end{figure}

Figure \ref{fig1} shows the results in Task3 and Task6 (with 6 and 10 categories respectively). It can be observed that the distributed ParMod is slightly superior to the standalone ParMod, with SAC as the baseline. The reward curves exhibit a basically consistent ascending trend, with PPO as the baseline. In a word, the distributed ParMod is more efficient than the standalone ParMod. The primary reason is that, excessive threads will lead to contention for computational resources in standalone ParMod, resulting in mutual blocking among threads. Distributed ParMod can mitigate this issue by distributing computational tasks across multiple machines. The throughput rate of the distributed ParMod is 241\% (218\% for task3 and 265\% for task6) higher than the standalone ParMod. Furthermore, we believe the performance of distributed ParMod will be more flexible in the case an NMT needs more categories of task phases.

\section{Summary and Discussion}

Learning an NMT is challenging since an agent receives reward for complex, temporally-extended behaviors sparsely. In this paper, we propose a novel modular framework ParMod to learn NMTs which are specified by a formal language LTL$_f$. We investigate an efficient parallel training approach based on a task division algorithm which relies on the automaton structure of LTL$_f$. We emphasize that LTL$_f$ does not deeply bind to our approach. Other temporal logics may also be applicable with slight change of reward functions. The experimental results on benchmark problems demonstrate the effectiveness and feasibility of our approach, showing faster training speed and convergence rate, better policy quality, and higher success rate, versus related studies. In addition, the scalability, parameter sensitivity, and the feasibility for distributed systems of ParMod have also been thoroughly evaluated.

Overall, as the first parallel framework designed for NMTs (to the best of our knowledge), we believe our approach supports a practical way to deal with non-Markovian property. So far, ParMod classifies task phases according to the static structure of DFA. We will take the environment state into account to join the classification process dynamically. This is a more precise way to measure the distance to the goal. We also plan to leverage the advantages of distributed RL and ParMod, in order to obtain more powerful capabilities to learn NMTs in future work.

\section*{Acknowledgements} 
The authors really appreciate for the reviewing efforts of the editors and the reviewers. This research is supported by National Natural Science Foundation of China (62372347,62192734,62202361,62172322,61806158); China Postdoctoral Science Foundation (2019T120881,2018M643585); Key Science and Technology Innovation Team of Shaanxi Province (2019TD-001); Fundamental Research Funds for the Central Universities (YJSJ24015); Innovation Fund of Xidian University (YJSJ24015).

	\bibliographystyle{model5-names}

	\bibliography{cas-refs}




\end{sloppypar}
\end{document}